\documentclass{article} 
\usepackage{iclr2027_arxiv,times}

\usepackage{amsmath,amsfonts,bm}

\def\eqref#1{equation~\ref{#1}}

\def\1{\bm{1}}

\DeclareMathAlphabet{\mathsfit}{\encodingdefault}{\sfdefault}{m}{sl}
\SetMathAlphabet{\mathsfit}{bold}{\encodingdefault}{\sfdefault}{bx}{n}

\usepackage{hyperref}
\hypersetup{hidelinks}
\usepackage{url}
\usepackage{booktabs}
\usepackage{amsmath}
\usepackage{amssymb}
\usepackage{amsthm}
\usepackage{mathtools}
\usepackage{microtype}
\usepackage{subcaption}
\usepackage{multirow}
\usepackage{xcolor}

\title{Structure--Residual Inductive Biases\\
for Language Model Pretraining}

\author{
Hai Huang$^{1}$ \quad
Bartosz \.{Z}uk$^{2}$ \quad
Rassul Magauin$^{3}$ \quad
Jiaxin Huang$^{3}$ \quad
Xue Liu$^{3}$ \\[0.6em]
{\normalfont $^{1}$Palona AI \quad
$^{2}$Institute of Computer Science, Polish Academy of Sciences} \\
{\normalfont $^{3}$Mohamed bin Zayed University of Artificial Intelligence}
}

\newtheorem{proposition}{Proposition}
\newtheorem{lemma}{Lemma}

\iclrpreprint
\begin{document}

\maketitle
\begin{abstract}
Next-token prediction is relied upon to learn both predictive
behavior and internal representations. We ask whether representation structure itself should be
an explicit inductive bias. Motivated by the two-part coding view of
Kolmogorov complexity, we study a structure--residual principle in which a
representation separates structure from realization detail.
We instantiate this principle in two complementary ways. Boundary Bottleneck
Concept (BBC) routes information from
token segments through boundary carriers while preserving local
autoregressive computation. Continuous Concept (CC) regularizes representation geometry
and induces a spectral decomposition into low-frequency
structure and high-frequency residual. We derive coding bounds for both views and show that multiplicity-corrected boundary attention becomes exact as the within-segment path energy vanishes.
Empirically, BBC and CC exhibit strikingly different depth preferences: BBC
works best in early layers, whereas CC is most effective in later layers.
At 300M training tokens, combining BBC and CC
improves NLL by 1.44\% on GPT-2 Medium. These gains persist with
training runs up to 20B tokens. Representation diagnostics show that BBC
increases the relative predictive importance of boundary states, while CC
reduces spectral residual energy. Together, these results suggest that
structure--residual biases can improve language-model
pretraining beyond next-token prediction. Code: \url{https://github.com/haihuang-ml/structure-residual}
\end{abstract}

\begin{figure*}[h]
    \centering

    \begin{minipage}[c]{0.39\textwidth}
        \centering
        \captionsetup[subfigure]{skip=0pt}
        \begin{subfigure}[t]{\linewidth}
            \centering
            \includegraphics[width=\linewidth]{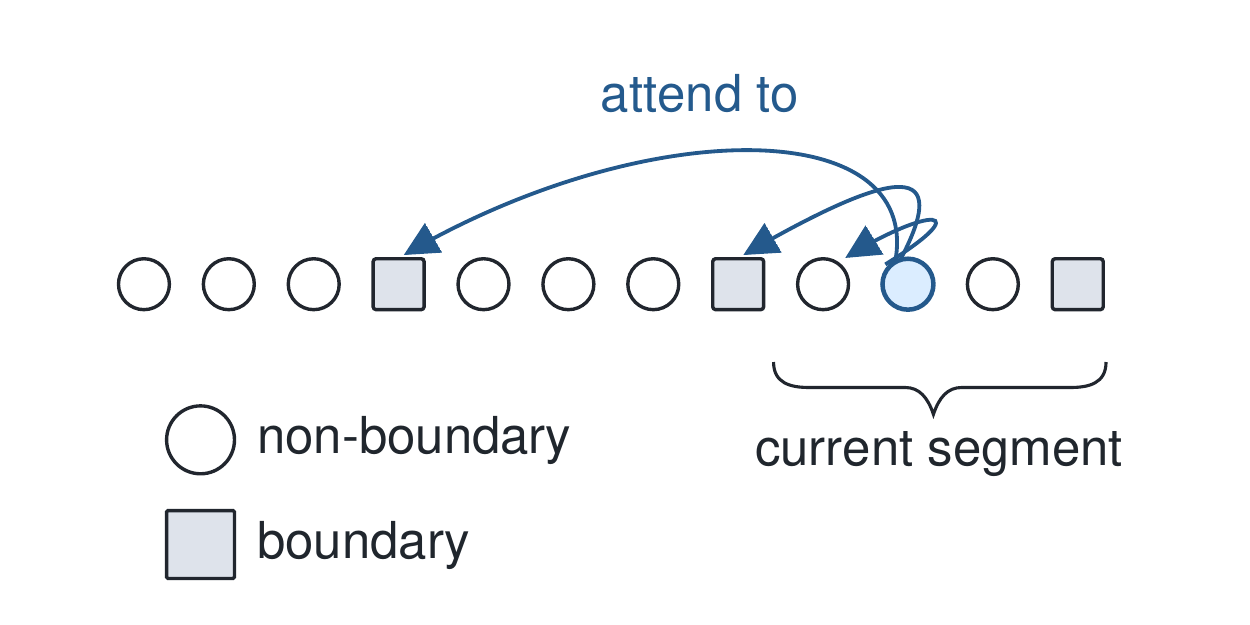}
            \caption{Boundary Bottleneck Concept}
            \label{fig:teaser_bbc}
        \end{subfigure}
        \begin{subfigure}[t]{\linewidth}
            \centering
            \includegraphics[width=\linewidth]{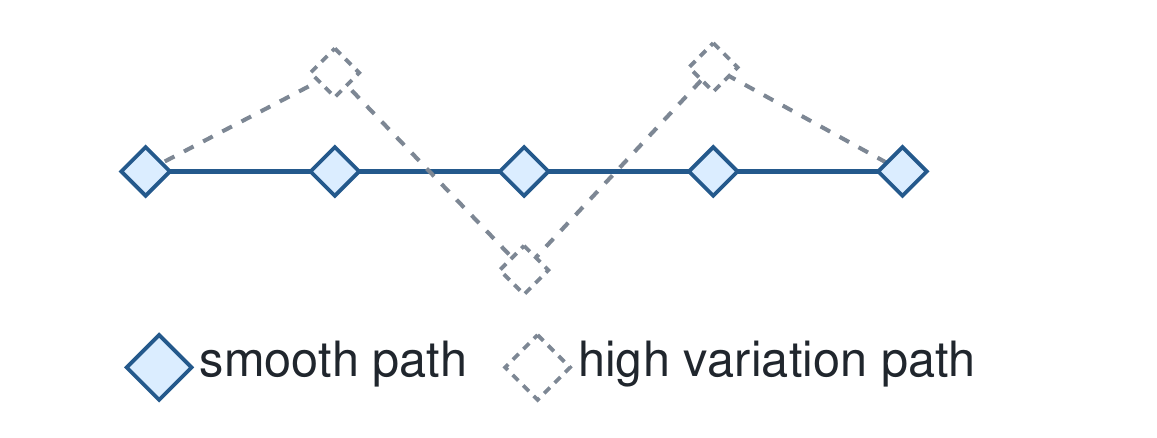}
            \caption{Continuous Concept}
            \label{fig:teaser_cc}
        \end{subfigure}

    \end{minipage}
    \hfill
    \begin{minipage}[c]{0.6\textwidth}
        \centering

        \begin{subfigure}[c]{\linewidth}
            \centering
            \includegraphics[width=\linewidth]{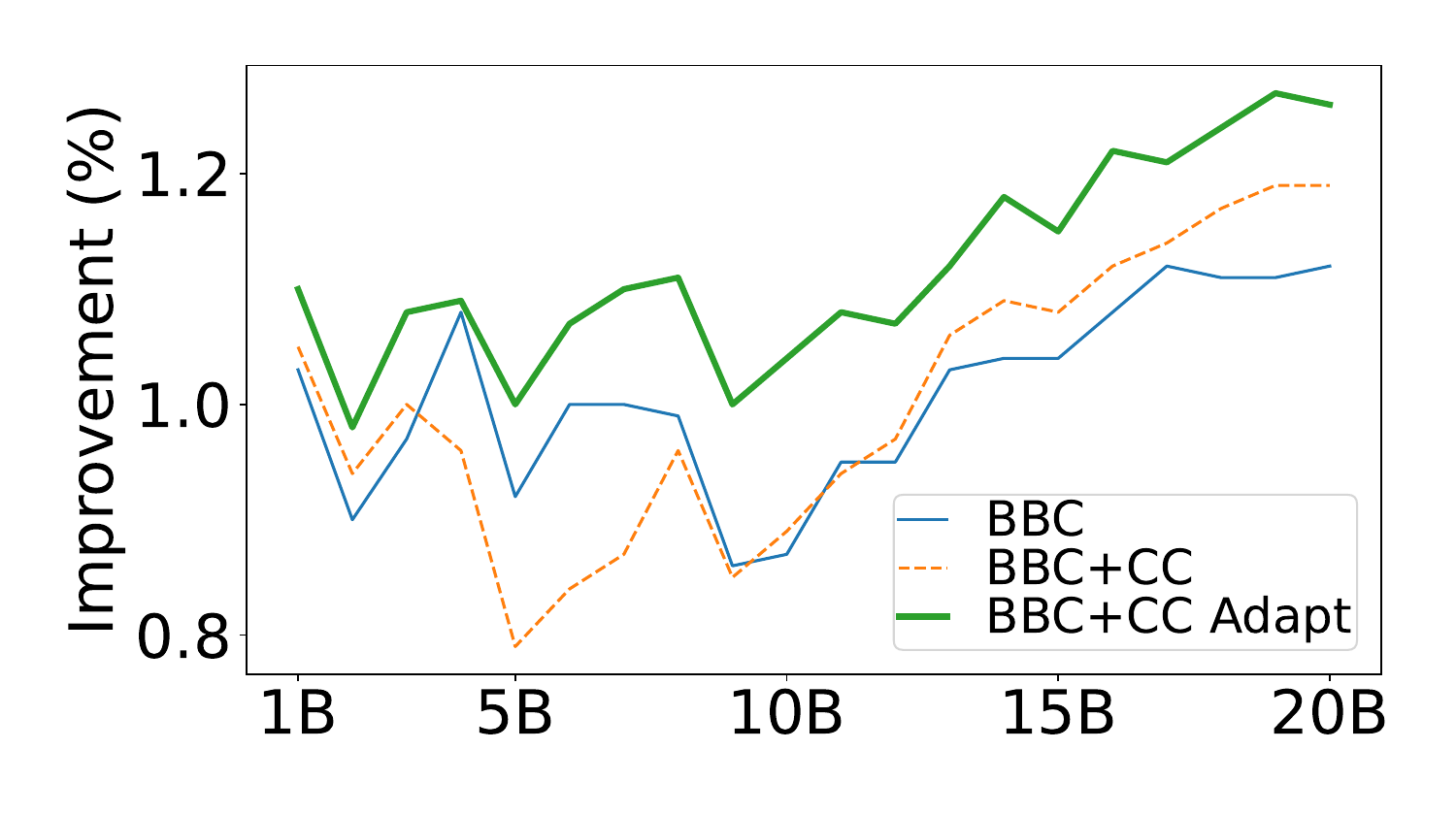}
            \caption{The gains persist through 20B training tokens.}
            \label{fig:teaser_scaling}
        \end{subfigure}

    \end{minipage}

    \caption{
        \textbf{Structure--residual inductive biases for language model pretraining.}
        (a) BBC restricts cross-segment communication to boundary carriers while
        retaining local autoregressive detail.
        (b) CC encourages smooth positional structure and reduces high-frequency
        residual variation.
        (c) The resulting improvements persist over long training horizons.
    }
    \label{fig:teaser}
\end{figure*}

\section{Introduction}

The dominant recipe for improving language models is remarkably simple: scale the model, the data, and the compute, and optimize next-token prediction (NTP)~\citep{kaplan2020scaling,hoffmann2022compute}. Under this paradigm, the training process---primarily gradient-based optimization---is trusted not only to learn the predictive task, but also to discover how the information required for that task should be internally represented.

We ask a different question: \emph{should representation structure itself be an inductive bias?}

A principled starting point is the two-part coding view of Kolmogorov complexity. An object can be described by first specifying a \textit{structure} $Z$, and then specifying the \textit{residual} information required to recover
the particular realization $X$:
\begin{equation*}
    \mathsf{K}(X)
    \leq
    \mathsf{K}(Z)
    +
    \mathsf{K}(X\mid Z)
    +
    O(1).
\end{equation*}
This suggests a general inductive principle for language models: encourage internal representations to discover a favorable balance between structure and residual. 

What should such structure look like in a language model? We draw inspiration from two basic functions involved in language modeling.

First, \textit{perception}
often involves abstraction. Phrases such as ``France's capital'' and ``the capital of France'' differ at the token level but share substantial predictive structure. This suggests compressing information that must be carried across spans into a compact representation. However, a structure-only representation is insufficient: the exact surface realization still matters for autoregressive prediction. 
Motivated by this structure--residual balance, we introduce \emph{Boundary Bottleneck Concept} (BBC), which routes information from a completed segment to later segments through a boundary carrier while preserving ordinary token-level autoregression within the current segment (Figure~\ref{fig:teaser_bbc}). Related approaches include MEGABYTE's
combination of global patch modeling and local autoregressive byte decoding
\citep{yu2023megabyte}, and gist tokens that compress prompt information
through attention bottlenecks \citep{mu2023gist}. BBC instead uses existing
segment-end states as carriers. 

Second, \textit{generation} requires fine positional information: every token position remains potentially relevant to the next-token distribution. This appears to leave little room for structural compression. Yet positional detail need not imply arbitrary representational complexity. A sequence of hidden states may preserve every token position while still admitting a simple underlying trajectory. Motivated by this observation, we study \emph{Continuous Concept} (CC), which
encourages neighboring hidden states to vary smoothly (Figure~\ref{fig:teaser_cc}). We analyze a spectral structure--residual decomposition: low-frequency modes form the structured component, while CC energy bounds the high-frequency residual and its finite-precision index length under a shared coding convention.


On GPT-2 Small, BBC improves evaluation NLL by approximately $0.99\%$ at 300M training tokens, while BBC+CC improves it by $1.30\%$. On a deeper GPT-2 Medium model, the gains are $1.11\%$ and $1.44\%$, respectively. BBC is not explained by sparsity alone: same-density sliding-window and random sparse-attention controls do not reproduce its gain, and randomizing the location of BBC carriers significantly degrades performance. Similarly, CC is not explained by either of its two components. Their improvements also persist with longer training, reaching $1.12\%$ for BBC and $1.26\%$ for BBC+CC through our 20B-token runs.

Our empirical study also reveals that the two biases favor different depths. BBC works best near
the bottom of the network, while CC is most effective at
later layers. 
Interestingly, the depth asymmetry is consistent with the broader intuition that different depths of a Transformer perform different computational roles: earlier representations can benefit from compact cross-span structure, whereas later representations retain fine token-level access while benefiting from simpler feature geometry.

Our results suggest that scaling NTP is not the only lever for improving language-model training. Explicit biases on \emph{how information is represented} can improve the same predictive objective. 

Our contributions are:
\begin{itemize}
    \item We use the two-part coding principle to analyze structure--residual inductive biases in language-model representations.

\item We introduce BBC, which compresses cross-segment communication through boundary carriers, and characterize CC as a spectral structure--residual bias with explicit reconstruction and residual-description bounds; their parameters $K$ and $\beta$ set the balance.

    \item We find that BBC and CC favor different depths, outperform matched architectural and regularization controls, and deliver gains that persist across model sizes and long training horizons; the $K$ and $\beta$ sweeps are U-shaped, consistent with this role.

    
    
\end{itemize}

\section{The Structure--Residual Coding Principle}
\label{sec:structure_residual}

Two-part coding describes an object through its structure and
the detail needed to identify its particular realization
\citep{vereshchagin2004structure}. We write $\mathsf{K}$ for prefix
Kolmogorov complexity. For finitely described objects,
\begin{equation*}
    \mathsf{K}(X\mid C)
    \leq
    \mathsf{K}(Z\mid C)
    +
    \mathsf{K}(X\mid Z,C)
    +
    O(1).
\end{equation*}
Here $Z$ describes structure and the conditional term describes
the remaining realization detail.

We use this principle to motivate concrete coding bounds.
Let $\Gamma$ denote a coding convention shared by the encoder
and decoder. Suppose that, after decoding $Z$, a residual
codebook with $M_\delta(Z)$ entries permits reconstruction of
$X$ to accuracy $\delta$. A two-part code then satisfies
\begin{equation*}
    \ell_\delta^{\mathrm{SR}}(X\mid C,\Gamma)
    \leq
    \ell_{\mathrm{str}}(Z\mid C,\Gamma)
    +
    \left\lceil\log_2 M_\delta(Z)\right\rceil.
\end{equation*}
The structure code is self-delimiting, and the decoded structure
and coding convention determine the residual index length.
Any codebook or parameter not shared in advance must also
be described.

For discrete objects, the residual codebook can enumerate
realizations sharing a structure. For continuous representations,
a covering codebook gives reconstruction at a chosen precision.

The bounds below concern these explicit codes; neither BBC nor CC directly optimizes Kolmogorov complexity.

BBC and CC implement different parts of this viewpoint. BBC restricts direct cross-segment communication while retaining local autoregressive detail; Section~\ref{sec:bbc} shows that its segment length $K$ sets this restriction and that both extreme values of $K$ give the vanilla model. CC controls the
residual outside a low-frequency representation of the hidden-state trajectory; Section~\ref{sec:cc} bounds this residual by the CC energy and relates the weight $\beta$ to the energy budget. Section~\ref{sec:cc_to_bbc} connects the two mechanisms, and Section~\ref{sec:experiments} locates the best $K$ and $\beta$ empirically.

\section{Boundary Bottleneck Concept}
\label{sec:bbc}

Boundary Bottleneck Concept (BBC) instantiates the structure--residual
principle through a granularity hyperparameter $K$. A sequence is divided
into segments of $K$ tokens. At a selected Transformer layer, each completed
segment communicates with later segments only through its boundary state,
while ordinary autoregressive computation is preserved inside the current
segment. 


\textbf{Boundary-limited attention.}\ \ \ For notational simplicity, assume that $K$ divides $N$ and
index segments by $i=0,\ldots,N/K-1$. Empty ranges in the
attention mask are interpreted as empty sets.

For fixed segment length $K$, define $S_i = \left(x_{iK+1},\ldots,x_{(i+1)K}\right)$. 
The final position of each segment serves as its \textit{boundary carrier}.

For a query token $x_t$ in segment $s=\left\lfloor\frac{t-1}{K}\right\rfloor$,
BBC replaces the ordinary causal attention pattern at the selected layer by
\begin{equation}
    \mathcal{A}_{\mathrm{BBC}}(t)
    =
    \{K,2K,\ldots,sK\}
    \cup
    \{sK+1,\ldots,t\}.
    \label{eq:bbc_mask}
\end{equation}
Once a segment is completed, its $K$ token states are no longer independently
accessible to later segments at the BBC layer. 
The current segment remains autoregressive. Let $C_s$ denote
the causal context preceding segment $S_s$, and define
$x_{s,r}=x_{sK+r+1}$. The segment distribution factorizes as
$p_\theta(S_s\mid C_s) = \prod_{r=0}^{K-1} p_\theta \left( x_{s,r}\mid C_s,x_{s,<r} \right)$.

All other layers retain the standard causal mask. Since other layers and the residual stream can retain
additional prefix information, this layer-local restriction
does not imply that the full model conditions only on boundary
carriers.

\textbf{Structure and realization residual.}\ \ \ Let $a=\phi(S)$ be a discrete structure assigned to a $K$-token
segment. Its realization class is $\mathcal{S}_K(a) = \{S'\in\mathcal{V}^K:\phi(S')=a\}$, where $\mathcal{V}$ is the vocabulary. 
The segment distribution induces a structure distribution:
$p_\theta(a\mid C_s) = \sum_{S'\in\mathcal{S}_K(a)} p_\theta(S'\mid C_s)$.

For an observed segment $S$ with $\phi(S)=a$ and positive
probability,
\begin{equation}
    -\log p_\theta(S\mid C_s)
    =
    -\log p_\theta(a\mid C_s)     
    -\log p_\theta(S\mid a,C_s).
    \label{eq:bbc_prob_residual}
\end{equation}
Thus exact segment prediction specifies both the structure
and its realization. In particular, $-\log p_\theta(a\mid C_s)\leq-\log p_\theta(S\mid C_s)$.
The realization class also gives a coding bound. Assume that
$\phi$ is computable and that the finite vocabulary, segment
length, and structure map are supplied to the decoder as
$\Gamma_{\mathrm{BBC}}$. The class can then be enumerated,
and a realization identified by its index:
$$\mathsf{K}(S\mid C_s,\Gamma_{\mathrm{BBC}})\leq\mathsf{K}(a\mid C_s,\Gamma_{\mathrm{BBC}})+\left\lceil\log_2|\mathcal{S}_K(a)|\right\rceil+O(1).$$
Since $\mathcal{S}_K(a)\subseteq\mathcal{V}^K$, the logarithm of the realization-class size satisfies
\begin{equation}
    \log_2|\mathcal{S}_K(a)|
    \leq K\log_2|\mathcal{V}|.
    \label{eq:bbc_residual_bound}
\end{equation}

\begin{proposition}[Segment length sets the bottleneck]
\label{prop:bbc_structure_residual}
Fix $N$ and the BBC layer. For a query at position $t\leq N$, write
$t=sK+r+1$ with $s=\lfloor (t-1)/K\rfloor$ and $0\leq r\leq K-1$.
\begin{enumerate}
    \item[(i)] \emph{Routing.} The query attends to the $s$ boundary
    carriers of the completed segments and to the $r+1$ positions
    $sK+1,\ldots,t$ of its current segment. Relative to the causal mask,
    it loses direct access to exactly $s(K-1)$ positions, the
    non-boundary tokens of the completed segments.
    \item[(ii)] \emph{Endpoints.}
    $\mathcal{A}_{\mathrm{BBC}}(t)=\{1,\ldots,t\}$ for every $t\leq N$
    if and only if $K=1$ or $K\geq N$. In both cases the BBC layer is a
    standard causal attention layer.
    \item[(iii)] \emph{Structure and realization.} The segment
    distribution admits the decomposition in
    Equation~\ref{eq:bbc_prob_residual}. Under the stated coding
    convention, a nonempty realization class requires at most
    $\lceil\log_2|\mathcal{S}_K(a)|\rceil\leq\lceil K\log_2|\mathcal{V}|\rceil$
    index bits.
\end{enumerate}
\end{proposition}

Parts (i) and (ii) hold for every $K\geq1$; the proof is in
Appendix~\ref{app:bbc_proofs}.

By (i), $K$ sets the split at the BBC layer: each completed segment is visible
there only through its boundary state, while up to $K$ positions of the
current segment remain directly visible. A key $\delta\geq1$ tokens back is
removed for a fraction $(\min(\delta,K)-1)/K$ of query offsets, so from
distance $K$ on exactly one key in $K$ remains: $K$ sets a distance scale, not
only a sparsity level (Lemma~\ref{lem:bbc_profile},
Appendix~\ref{app:bbc_theory}). By (ii), $K=1$ and $K\geq N$ give the vanilla
model.

The probability identities hold for any segment distribution;
BBC contributes the architectural routing restriction.
Its structure--residual interpretation is therefore a compact
cross-segment carrier together with retained local
autoregressive computation. The boundary need not equal
$\phi(S)$ or be predictively sufficient by construction.

\section{Continuous Concept}
\label{sec:cc}

Continuous Concept (CC) is an optimization bias on the geometry of the
hidden-state trajectory.
Let $N\geq2$, and let $H\in\mathbb{R}^{N\times d}$ have row $i$ equal to $h_i^\top$. The states are taken at a selected layer. CC penalizes adjacent variation:
\[\mathcal{L}_{\mathrm{CC}} = \frac{1}{N-1}\sum_{i=1}^{N-1}\|h_{i+1}-h_i\|_2^2.\]
The objective is $\mathcal{L}=\mathcal{L}_{\mathrm{NTP}}+\beta\mathcal{L}_{\mathrm{CC}}$, where $\beta\geq0$ controls the regularization strength.
In our implementation, $\mathcal{L}_{\mathrm{CC}}$ is also averaged over the
$d$ hidden coordinates, so each reported $\beta$ equals $d$ times the
coefficient defined here; the same holds for $\mathcal{L}_{\mathrm{scale}}$ and
$\mathcal{L}_{\mathrm{CC\text{-}2}}$ in Section~\ref{sec:experiments}.

We analyze the trajectory as low-frequency structure plus a high-frequency residual. The CC objective controls an upper bound on the residual magnitude.

\textbf{Spectral structure and residual.}\ \ \ Let $L$ be the unit-weight path Laplacian. The unnormalized
CC energy is
$E(H)=\sum_{i=1}^{N-1}\|h_{i+1}-h_i\|_2^2=\operatorname{Tr}(H^\top L H)=(N-1)\mathcal{L}_{\mathrm{CC}}$.
Write $L=Q\Lambda Q^\top,\ 0=\lambda_1<\lambda_2\leq\cdots\leq\lambda_N$,
with orthonormal eigenvectors. These form a graph-frequency
basis over token positions \citep{shuman2013graphs}. This is a vector-valued path instance of the graph-smoothness objective
underlying Laplacian Eigenmaps \citep{belkin2003laplacian}.

For $1\leq m<N$, define $Q_m=[q_1,\ldots,q_m],\ H_m=Q_mQ_m^\top H$. $H_m$ is the low-frequency structure, and $R_m=H-H_m$ is the spectral residual.

\begin{proposition}[Spectral residual bound]
\label{prop:cc_residual}
For $1\leq m<N,\ \|R_m\|_F^2 \leq \frac{E(H)}{\lambda_{m+1}}$.
\end{proposition}

\begin{proof}
Let $\widehat H=Q^\top H$. Orthogonality gives $\|R_m\|_F^2=\sum_{r>m}\|\widehat H_r\|_2^2,\ E(H)=\sum_r\lambda_r\|\widehat H_r\|_2^2$.
Therefore
$\|R_m\|_F^2\leq\frac{1}{\lambda_{m+1}}\sum_{r>m}\lambda_r\|\widehat H_r\|_2^2\leq\frac{E(H)}{\lambda_{m+1}}$.
\end{proof}

For a chosen cutoff, the approximation is accurate when
$E(H)/\lambda_{m+1}$ is small. 

\textbf{Residual codebook size.}\ \ \ Fix the dimensions, cutoff, and an energy budget $E_\star\geq0$.
For trajectories with $E(H)\leq E_\star$, the residual lies
in a subspace of dimension $D=(N-m)d$ and in a ball of radius $r_\star=\sqrt{\frac{E_\star}{\lambda_{m+1}}}$.
A $D$-dimensional Euclidean ball of radius $r_\star$ admits
a cover by at most 
\begin{equation}
    \left(1+\frac{2r_\star}{\eta}\right)^D
\label{eq:cc_covering_number}
\end{equation}
balls of radius $\eta>0$ \citep{vershynin2018hdp}.

\begin{proposition}[Spectral residual index length]
\label{prop:cc_codebook}
For a shared covering codebook satisfying
Equation~\ref{eq:cc_covering_number}, the bit length of an index
identifying a residual approximation with Frobenius error at most
$\eta$ satisfies
\[
L_\eta(R_m\mid\Gamma)
\leq
\left\lceil
(N-m)d\log_2\left(
1+\frac{2}{\eta}\sqrt{\frac{E_\star}{\lambda_{m+1}}}
\right)
\right\rceil.
\]
\end{proposition}

\begin{proof}
Choose a covering center within distance $\eta$ of the residual and transmit its index; Equation~\ref{eq:cc_covering_number} then gives the bound.
\end{proof}


The coding convention $\Gamma$ includes the dimensions, basis, cutoff,
energy budget, precision, and codebook; any unshared components incur
additional description cost. Allocating reconstruction error $\delta/2$ to
both structure and residual gives $\|H-(\widetilde H_m+\widetilde R_m)\|_F\leq\delta$, and hence
\begin{equation*}
    \ell_\delta^{\mathrm{SR}}(H\mid\Gamma)
    \leq
    \ell_{\delta/2}^{\mathrm{str}}(H_m\mid\Gamma)
    +
    \left\lceil
    (N-m)d\log_2\left(1+\frac{4r_\star}{\delta}\right)
    \right\rceil.
\end{equation*}
Thus, at fixed dimensions, cutoff, and precision, a smaller energy budget
reduces the residual description bound, while the structure retains its own
description cost.
A larger $\beta$ never increases the CC energy of the optimal model (Appendix~\ref{app:cc_beta}). Its strength at layer $l$ is $\kappa_l=\beta\rho_l$, where $\rho_l$ is the ratio of the CC to the NTP gradient (Appendix~\ref{app:cc_theory}).

\section{From spectral smoothness to boundary representation}
\label{sec:cc_to_bbc}

The structure--residual coding viewpoint motivates both interventions.
We now give a qualitative connection: low within-segment path energy
bounds the error of multiplicity-corrected boundary attention (Appendix~\ref{app:prop4_check} tests it on trained layers). The
reverse direction requires additional conditions; see
Appendix~\ref{app:bbc_cc_connection}.

Let $I_j$ contain the positions
of a completed $K$-token segment, with boundary
$b_j=\max I_j$, and define
$E_j=\sum_{\substack{t\in I_j\\t<b_j}}\|h_{t+1}-h_t\|_2^2$. 

\begin{lemma}[Boundary approximation under low path energy]
\label{lem:boundary_distance}
\begin{equation*}
    \max_{i\in I_j}\|h_i-h_{b_j}\|_2
    \leq\sqrt{(K-1)E_j}.
\end{equation*}
\end{lemma}

\begin{proof}
For $r=b_j-i, \ h_{b_j}-h_i=\sum_{t=i}^{b_j-1}(h_{t+1}-h_t)$. Hence
$\|h_{b_j}-h_i\|_2^2\leq r\sum_{t=i}^{b_j-1}\|h_{t+1}-h_t\|_2^2\leq(K-1)E_j$.
Equal, aligned increments attain the bound.
\end{proof}

Consider one attention head \citep{vaswani2017attention} with fixed query $q$,
shared linear key and value maps, and no attention dropout: $s_i=\frac{q^\top W_Kh_i}{\sqrt{d_k}},\ v_i=W_Vh_i$.
Partition all allowed key positions into groups $G_j$.
Completed segments form groups of size $K$; retained
local-prefix positions remain singleton groups.
Write $\kappa_j=|G_j|$ and let $b_j$ be the boundary
or singleton representative.

Define $g_j=\log\sum_{i\in G_j}e^{s_i}, \ u_j=\sum_{i\in G_j}\frac{e^{s_i}}{\sum_{r\in G_j}e^{s_r}}v_i$.
Regrouping the numerator and denominator gives $o=\operatorname{Attn}_{\mathrm{dense}}(q)=\sum_j a_j u_j, \ a=\operatorname{softmax}(g)$.
The representative approximation is $\widetilde g_j=s_{b_j}+\log\kappa_j, \ \widetilde u_j=v_{b_j}$,
with
$\widetilde o=\sum_j\widetilde a_j\widetilde u_j$
and $\widetilde a=\operatorname{softmax}(\widetilde g)$.
As in proportional attention \citep{bolya2023tome},
$\log\kappa_j$ preserves the mass of identical entries. Singleton groups are unchanged.

Let $c_K=\frac{\|q\|_2\|W_K\|_{\mathrm{op}}}{\sqrt{d_k}}, \ c_V=\|W_V\|_{\mathrm{op}}$, we have
$|g_j-\widetilde g_j|\leq c_K\sqrt{(K-1)E_j}, \ \|u_j-\widetilde u_j\|_2 \leq c_V\sqrt{(K-1)E_j}$ for each completed segments.
Indeed, each token score differs from its boundary score
by at most $c_K\sqrt{(K-1)E_j}$; bounding the sum of their
exponentials gives the first inequality. The second follows
because $u_j$ is a convex combination of values individually
close to $v_{b_j}$.

\begin{proposition}[Multiplicity-corrected boundary attention]
\label{prop:boundary_attention}
Let $\Delta_s=\max_j|g_j-\widetilde g_j|,\ \Delta_v=\max_j\|u_j-\widetilde u_j\|_2$.
If $\|\widetilde u_j\|_2\leq V_{\max}$ for every group,
including local-prefix singletons, then
\begin{equation*}
    \|o-\widetilde o\|_2
    \leq
    \Delta_v+2V_{\max}\tanh(\Delta_s).
\end{equation*}
\end{proposition}

\begin{proof}
The score bound implies
$e^{-2\Delta_s}\leq a_j/\widetilde a_j\leq e^{2\Delta_s}$.
Consequently, $|a_j-\widetilde a_j|\leq \tanh(\Delta_s)(a_j+\widetilde a_j)$,
so $\|a-\widetilde a\|_1\leq2\tanh(\Delta_s)$.
Therefore
$\|o-\widetilde o\|_2 \leq \sum_j a_j\|u_j-\widetilde u_j\|_2 + \sum_j|a_j-\widetilde a_j|\,\|\widetilde u_j\|_2 \leq \Delta_v+2V_{\max}\tanh(\Delta_s)$.
\end{proof}

For fixed $K$ and uniformly bounded sensitivity and value constants, the error vanishes as the maximum segment energy vanishes. This is a same-layer result for multiplicity-corrected attention, not directly for the uncorrected BBC mask. The bound assumes that $h_i$ enters the stated key and value maps; intervening normalization must be included in the sensitivity constants, and position-dependent maps or logit biases require additional terms. Appendix~\ref{app:rope} gives these terms for rotary position embeddings, for this result and for Proposition~\ref{prop:bbc_smoothness}; our GPT-2 models use
learned absolute position embeddings and need no such term.

\section{Experiments}
\label{sec:experiments}

\textbf{Setup.}\ \ \ 
We pre-train \texttt{gpt2-small} and \texttt{gpt2-medium} models \citep{radford2019language} with context length $N=1024$ on a randomly sampled 1B-token subset of \texttt{FineWeb-Edu} \citep{penedo2024fineweb}. We evaluate on a 1M-token holdout set sampled from the same dataset and report holdout NLL. The primary runs use a cosine learning-rate schedule over 1B tokens with a 5M-token linear warmup. To reduce compute, hyperparameter sweeps and ablations are evaluated after 100M or 300M training tokens while retaining the same 1B-token learning-rate decay schedule. Unless otherwise noted, we use three seeds, $\{82,23,37\}$, and report mean and standard deviation. For the 20B-token runs, we enlarge the holdout set to 10M tokens and use a single seed (82) to reduce computational cost.

We first tune the vanilla baseline by sweeping the learning rate
$\eta\in\{3\times10^{-4},6\times10^{-4},10^{-3},1.5\times10^{-3}\}$.
We then fix $\eta$ and separately sweep the BBC segment length
$K\in\{4,8,16,32,64\}$ and the CC regularization strength
$\beta\in\{0.005,0.0025,0.00125,0.000625\}$.
We do not retune $\eta$ for BBC or CC, avoiding method-specific learning-rate tuning.
We use $\eta=10^{-3}$ for \texttt{gpt2-small} and
$\eta=6\times10^{-4}$ for \texttt{gpt2-medium}.

\begin{figure*}[h]
    \centering

    \begin{subfigure}[t]{0.49\textwidth}
        \centering
        \includegraphics[width=\linewidth]{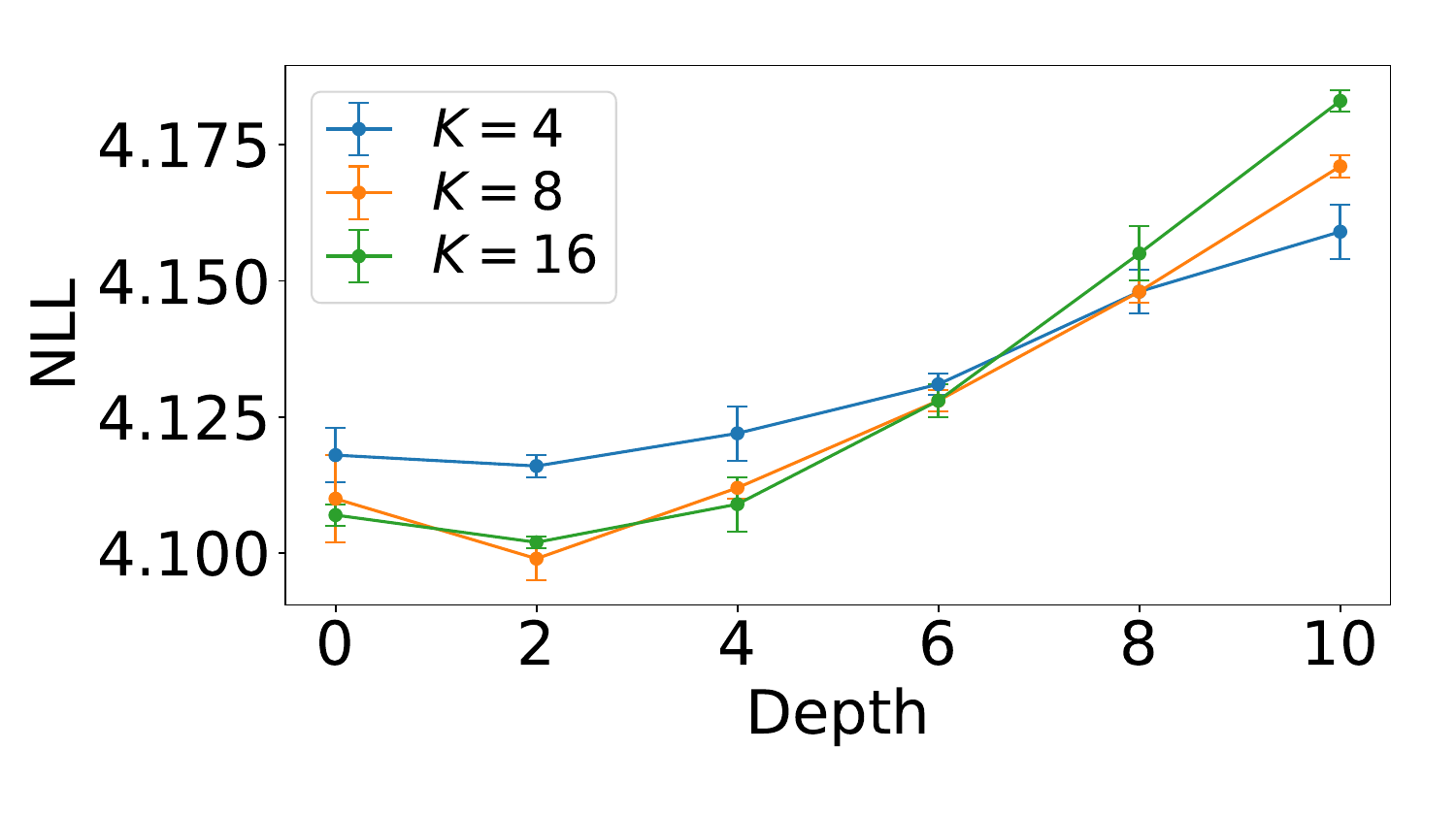}
        \caption{BBC on \texttt{gpt2-small}.}
        \label{fig:bbc_depth_small}
    \end{subfigure}
    \hfill
    \begin{subfigure}[t]{0.49\textwidth}
        \centering
        \includegraphics[width=\linewidth]{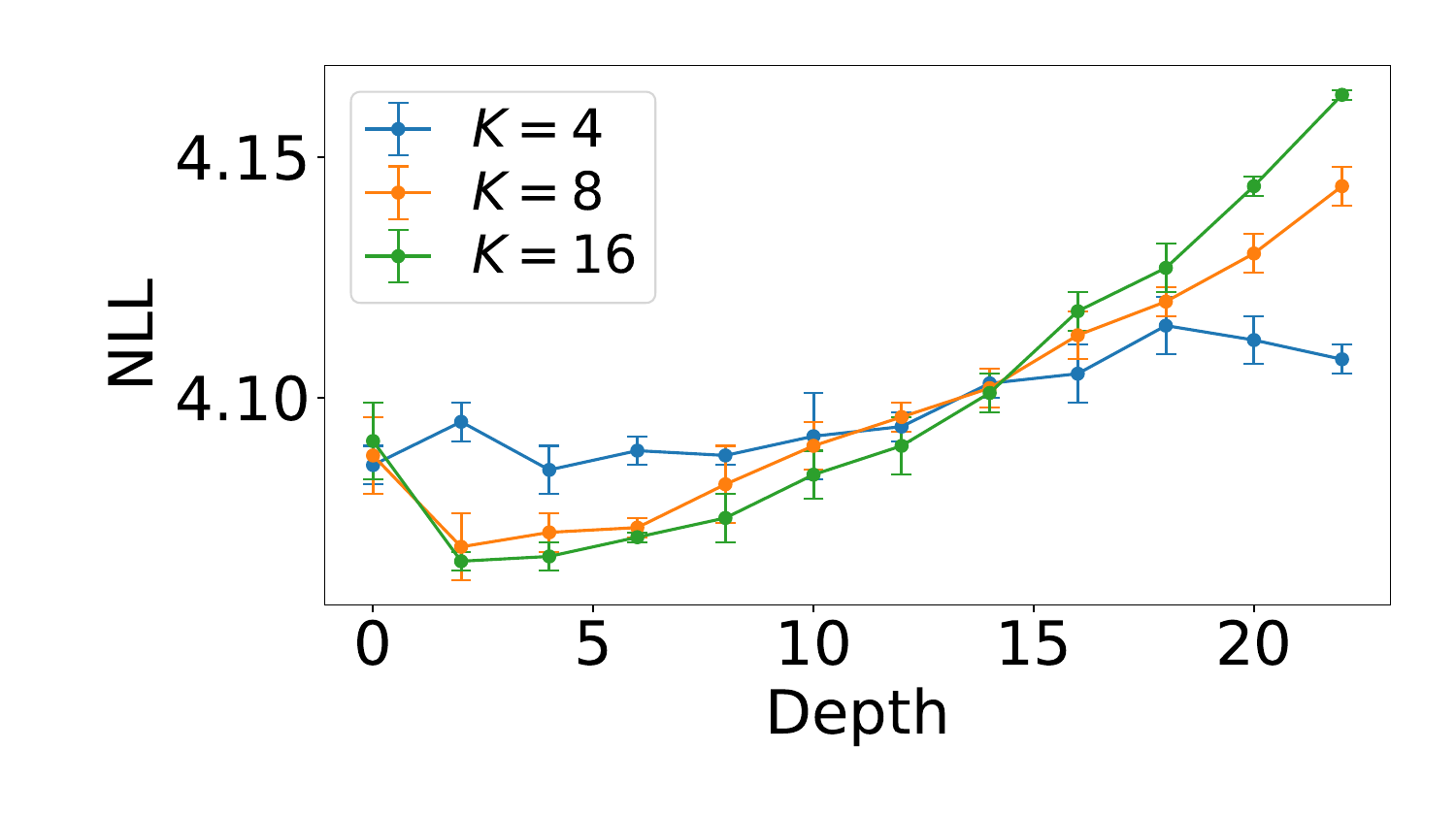}
        \caption{BBC on \texttt{gpt2-medium}.}
        \label{fig:bbc_depth_medium}
    \end{subfigure}
    \begin{subfigure}[t]{0.49\textwidth}
        \centering
        \includegraphics[width=\linewidth]{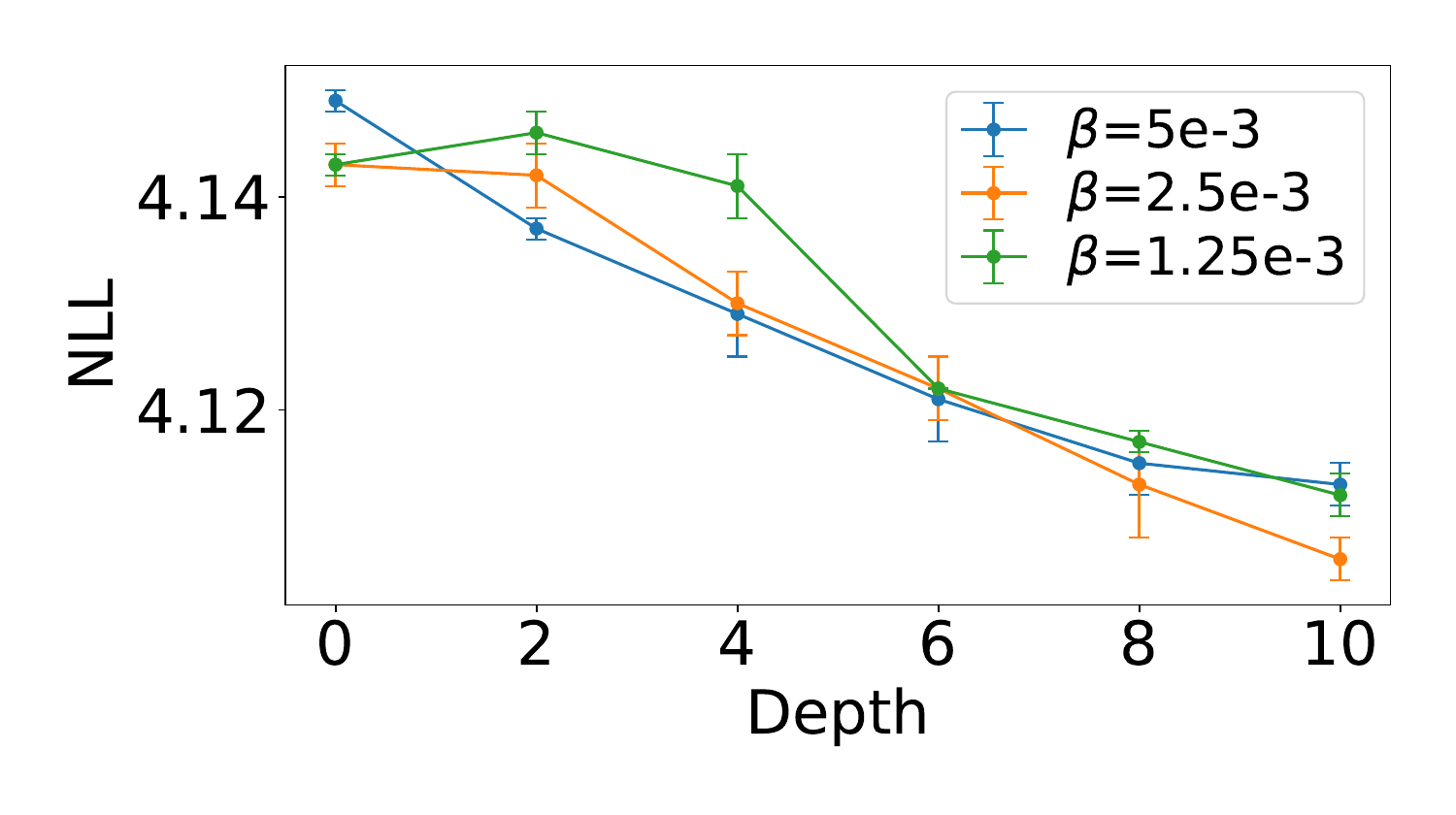}
        \caption{CC on \texttt{gpt2-small}.}
        \label{fig:cc_depth_small}
    \end{subfigure}
    \hfill
    \begin{subfigure}[t]{0.49\textwidth}
        \centering
        \includegraphics[width=\linewidth]{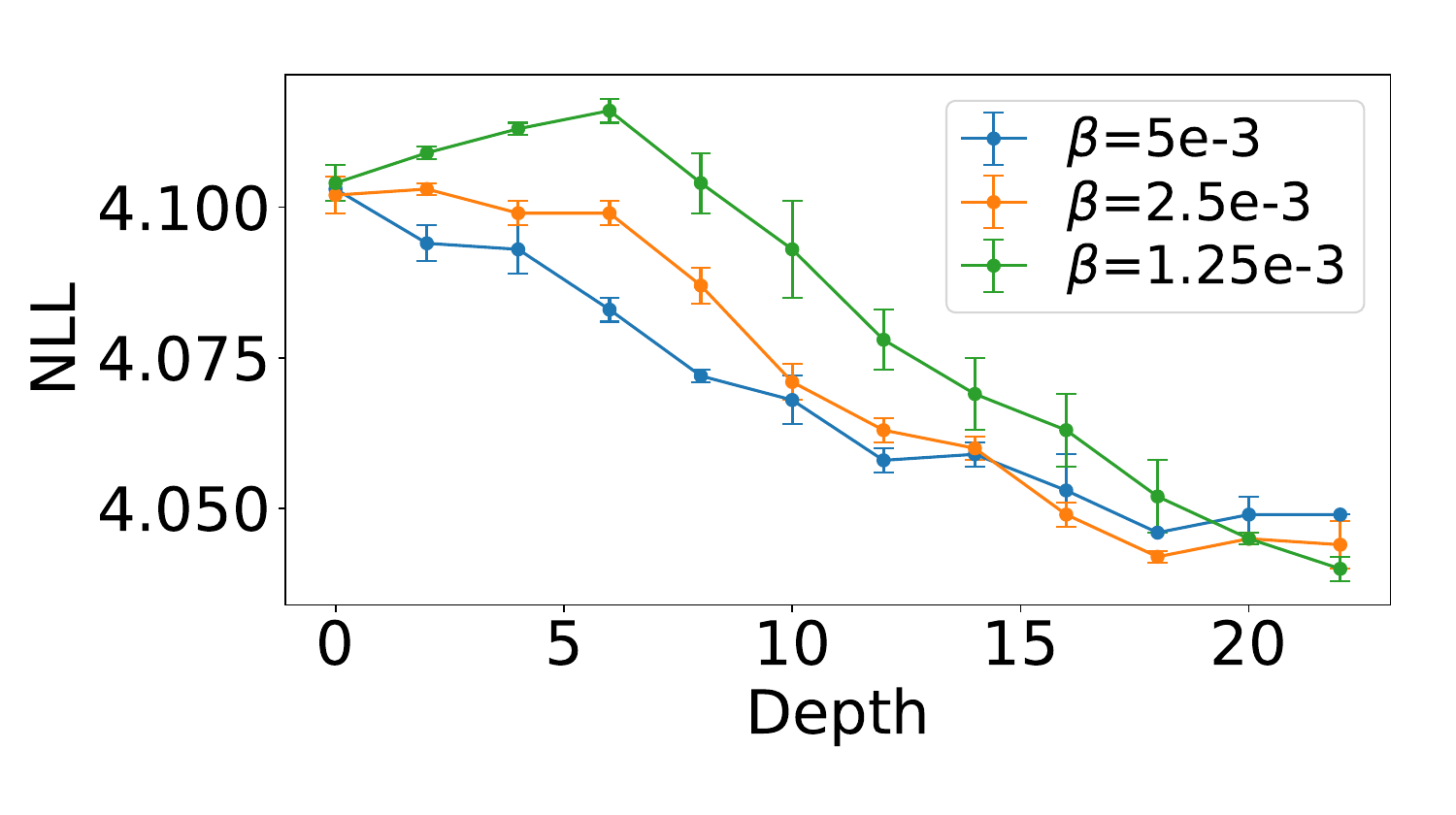}
        \caption{CC on \texttt{gpt2-medium}.}
        \label{fig:cc_depth_medium}
    \end{subfigure}
    \caption{
        \textbf{BBC and CC exhibit opposite depth preferences.}
        BBC is most effective at lower layers, while CC becomes more effective toward higher layers, consistently across both model sizes.
    }
    \label{fig:depth_sweeps}
\end{figure*}

\textbf{BBC and CC are depth-dependent.}\ \ \ 
Our first finding is a clear depth dependence for both methods. We sweep the intervention layer over the even-numbered Transformer layers for both \texttt{gpt2-small} and \texttt{gpt2-medium}, using $K\in\{4,8,16\}$ for BBC and $\beta\in\{0.00125,0.0025,0.005\}$ for CC (Figure~\ref{fig:depth_sweeps} and also Table~\ref{bbc-layers},~\ref{cc-layers} in Appendix~\ref{app:bbc-cc-layers}). The same qualitative pattern appears across both model sizes: BBC is most effective when applied at lower layers, whereas CC becomes more effective toward higher layers. This depth asymmetry was not built into either method. Interestingly, it is consistent with the motivating intuition that BBC is perception-inspired while CC is generation-inspired, suggesting that different Transformer depths may favor different representational functions.

In the main BBC/CC comparisons, BBC is applied at layer 2, while CC is applied at layer 10 for \texttt{gpt2-small} and layer 22 for \texttt{gpt2-medium}.

\begin{figure*}[h]
    \centering
    \begin{subfigure}[t]{0.49\textwidth}
        \centering
        \includegraphics[width=\linewidth]{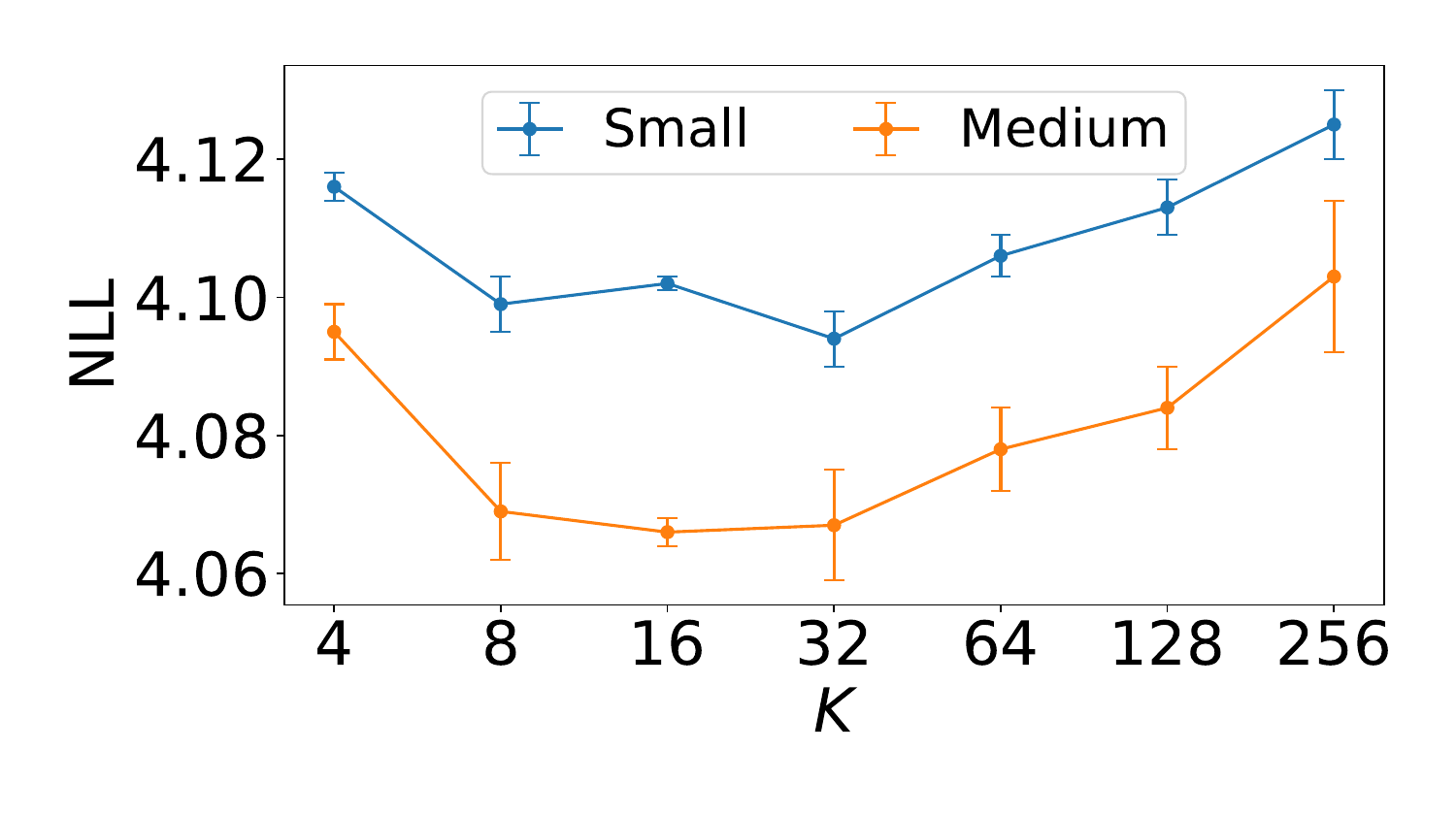}
        \caption{$K$ sweep at layer 2.}
        \label{fig:convex_k}
    \end{subfigure}
    \hfill
    \begin{subfigure}[t]{0.49\textwidth}
        \centering
        \includegraphics[width=\linewidth]{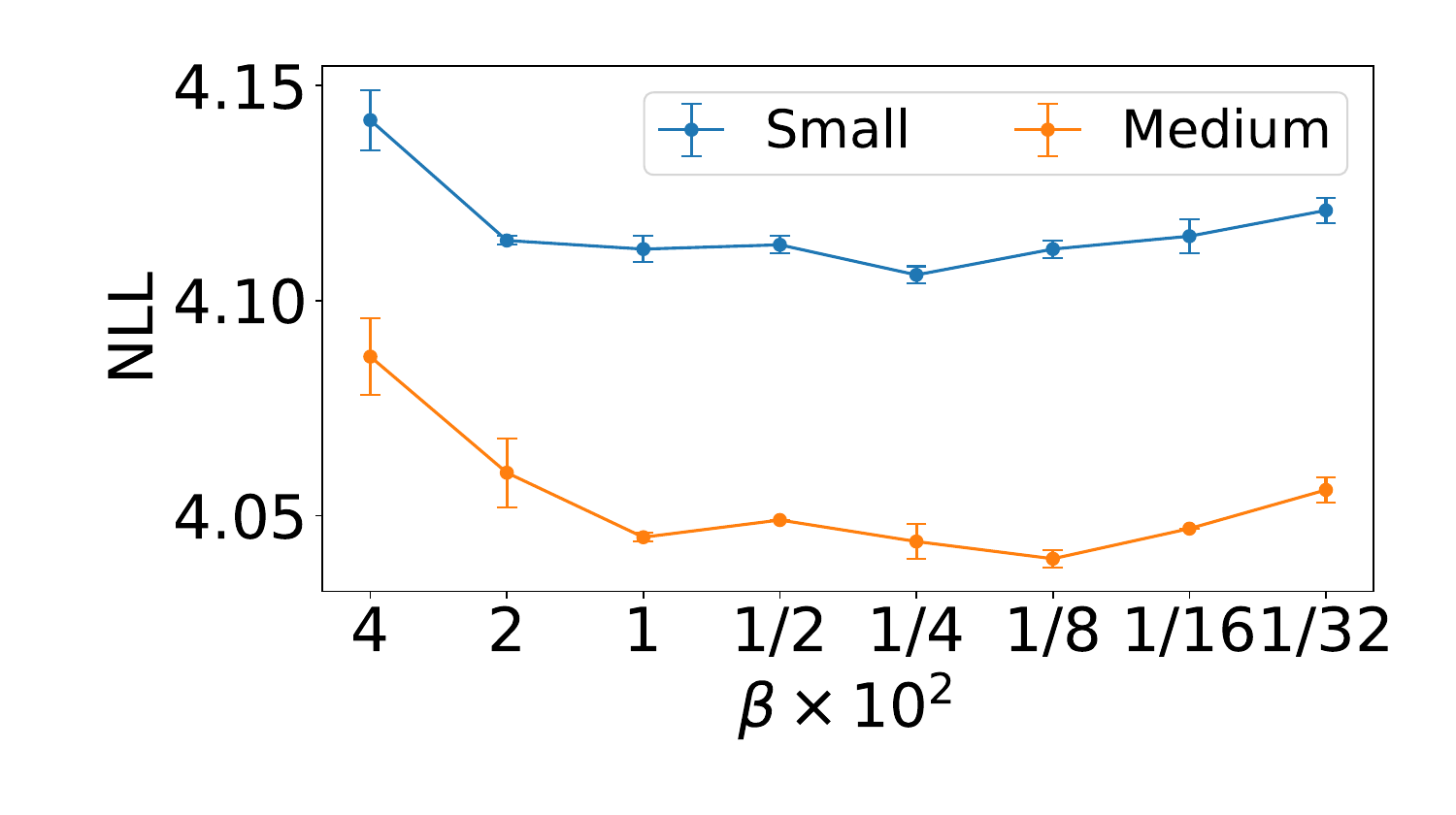}
        \caption{$\beta$ sweep at layer 10 for Small and 22 for Medium.}
        \label{fig:convex_beta}
    \end{subfigure}

    \caption{\textbf{Optimal $K$ and $\beta$.}
Both BBC and CC exhibit approximately U-shaped hyperparameter sweeps, indicating intermediate optima for $K$ and $\beta$. We sweep $K$ at layer 2 for both models, and $\beta$ at layers 10 and 22 for \texttt{gpt2-small} and \texttt{gpt2-medium}, respectively.
    }
    \label{fig:convex}
\end{figure*}

\textbf{Optimal $K$ and $\beta$.}\ \ \ 
We sweep $K$ and $\beta$ over broad ranges to identify their optimal values.
As shown in Figure~\ref{fig:convex}, both exhibit approximately U-shaped
performance curves with relatively broad optima: $K\in[8,32]$ and
$\beta\in[0.00125,0.0025]$ (Tables~\ref{bbc-layers}--\ref{tab:cc_beta_sweep}). For $\beta$, the best strength $\beta^\star\rho_l$ (Section~\ref{sec:cc}),
fitted on medium layers 10--20, predicts $\beta^\star=0.0013$ at layer
22 for GPT2 Medium and $0.0021$ at layer 10 for GPT2 Small; the sweeps find $0.00125$ and $0.0025$
(Table~\ref{tab:cc_strength}).

\textbf{300M significance tests and 1B persistence.}\ \ \ 
We find that the best $K$ and $\beta$ selected at 100M tokens retain their gains through 1B training tokens. For \texttt{gpt2-small}, we use $K=8$ and $\beta=0.0025$. For \texttt{gpt2-medium}, we use $K=16$ and $\beta=0.00125$.

At 300M tokens, the ordering is consistent across both model sizes: CC significantly outperforms vanilla, BBC significantly outperforms CC, and BBC+CC significantly outperforms BBC (Figure~\ref{fig:300m_nll} and also Table~\ref{small-300m-1b},~\ref{medium-300m-1b} in Appendix~\ref{app:300m-1b} for $p$-values).

\begin{figure*}[h]
    \centering
    \begin{subfigure}[t]{0.49\textwidth}
        \centering
        \includegraphics[width=\linewidth]{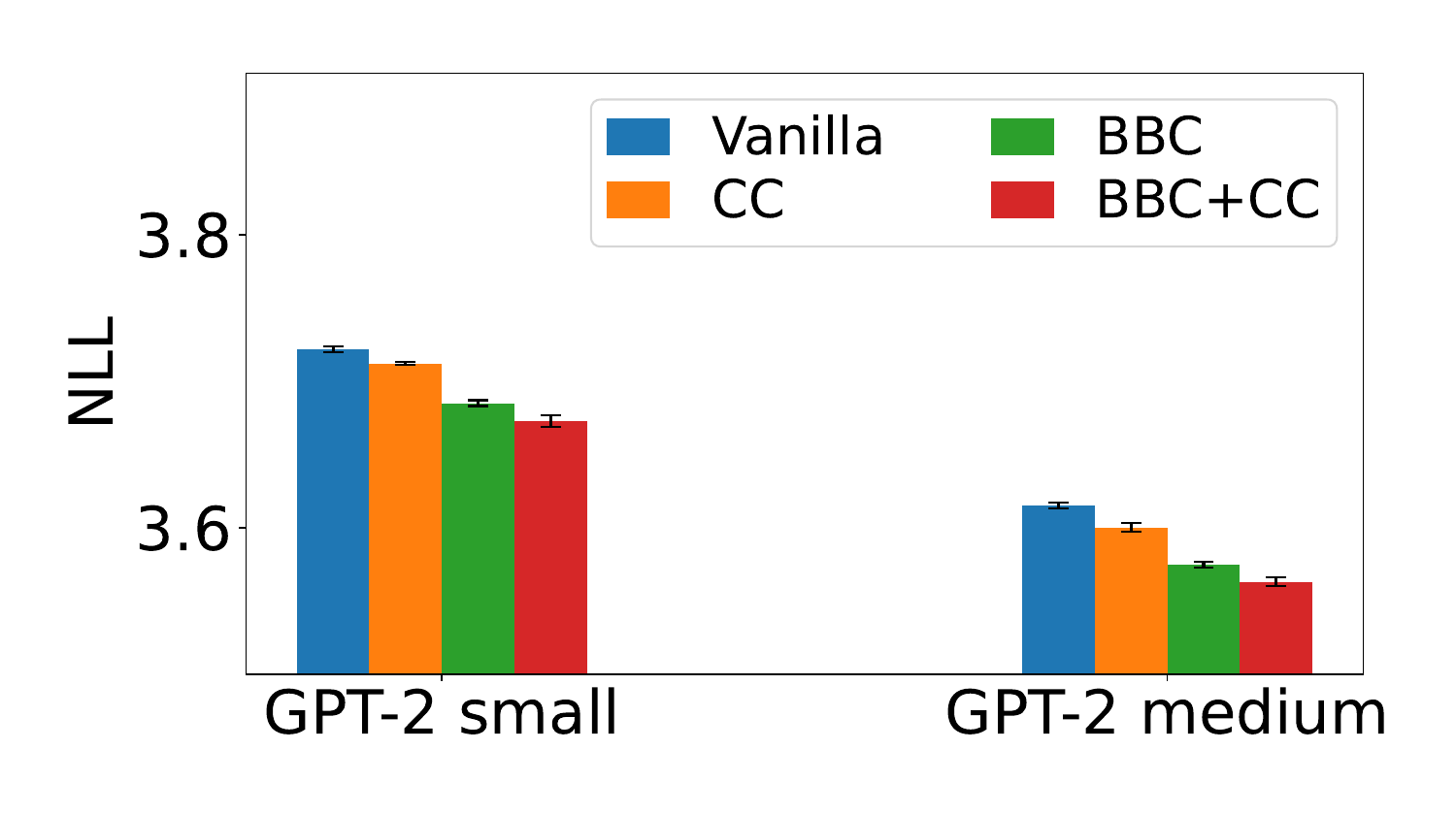}
        \caption{NLL at 300M training tokens.}
        \label{fig:300m_nll}
    \end{subfigure}
    \hfill
    \begin{subfigure}[t]{0.49\textwidth}
        \centering
        \includegraphics[width=\linewidth]{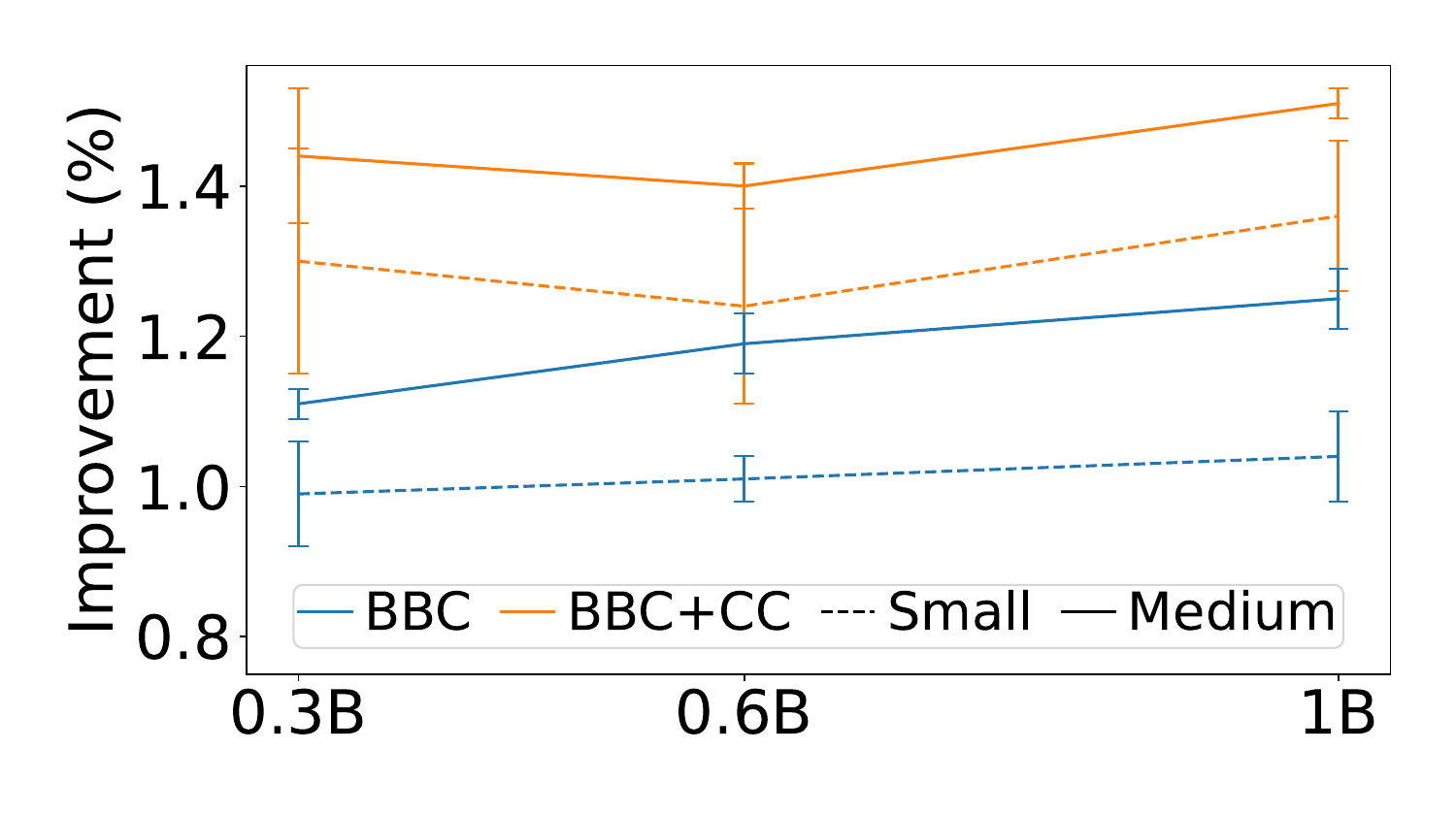}
        \caption{BBC and BBC+CC improvement through 1B training tokens.}
        \label{fig:bbc_cc_improvement}
    \end{subfigure}

    \caption{
        \textbf{Statistically significant gains at 300M tokens persist through 1B tokens.}
        (a) CC, BBC, and BBC+CC progressively improve over the vanilla baseline at 300M tokens.
        (b) BBC continues to outperforms vanilla as training proceeds, while BBC+CC maintains an additional margin over BBC.
    }
    \label{fig:significance_persistence}
\end{figure*}

The improvement persists as training continues. As shown in Figure~\ref{fig:bbc_cc_improvement}, from 300M to 1B tokens, BBC shows a steady upward trend in relative improvement over vanilla for both \texttt{gpt2-small} and \texttt{gpt2-medium}, while BBC+CC continues to maintain a statistically significant margin over BBC.

\begin{figure*}[h]
    \centering
    \begin{subfigure}[t]{0.48\textwidth}
        \centering
        \includegraphics[width=\linewidth]{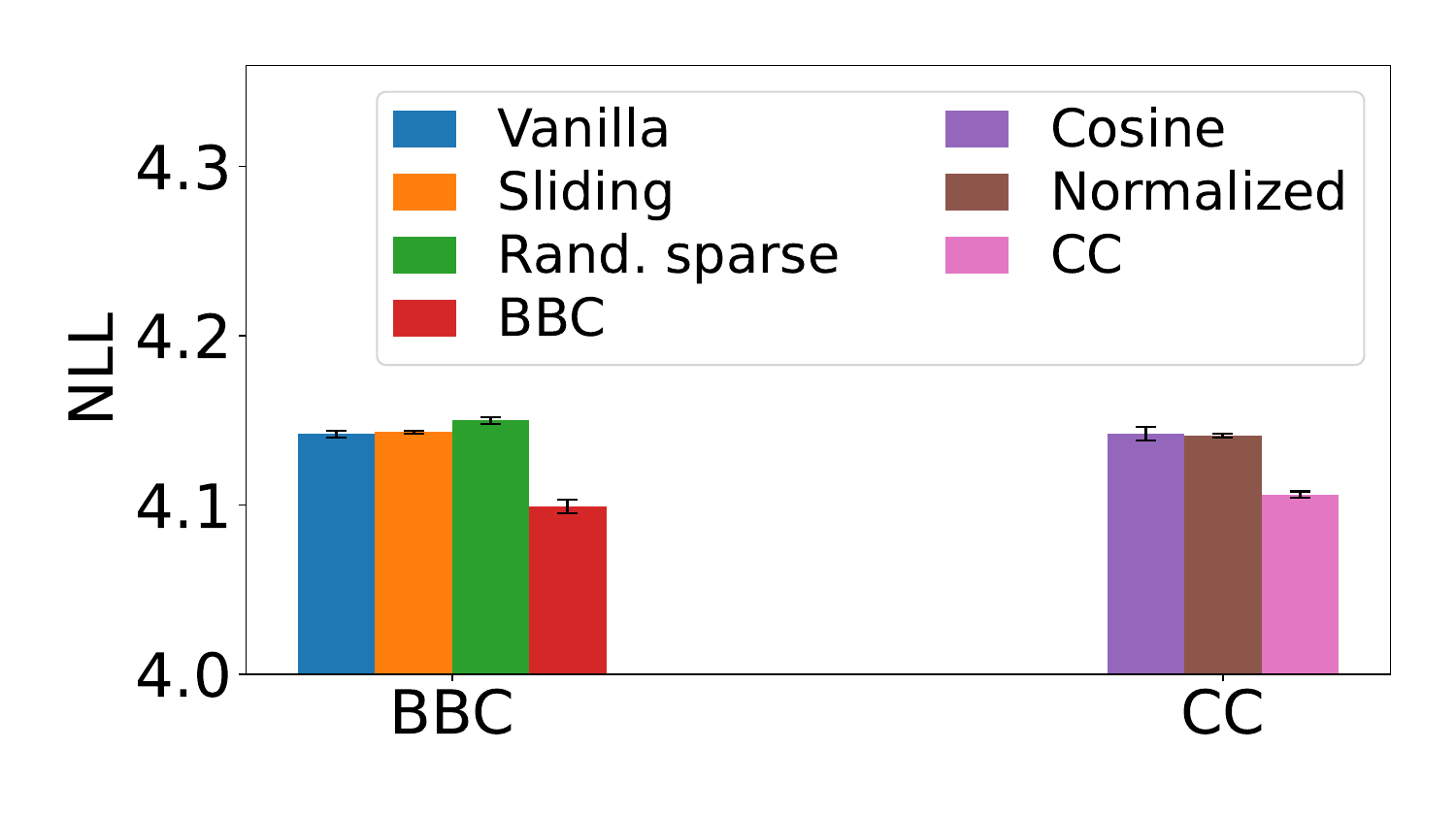}
        \caption{Ablations at 100M training tokens.}
        \label{fig:abl_100m}
    \end{subfigure}
    \hfill
    \begin{subfigure}[t]{0.48\textwidth}
        \centering
        \includegraphics[width=\linewidth]{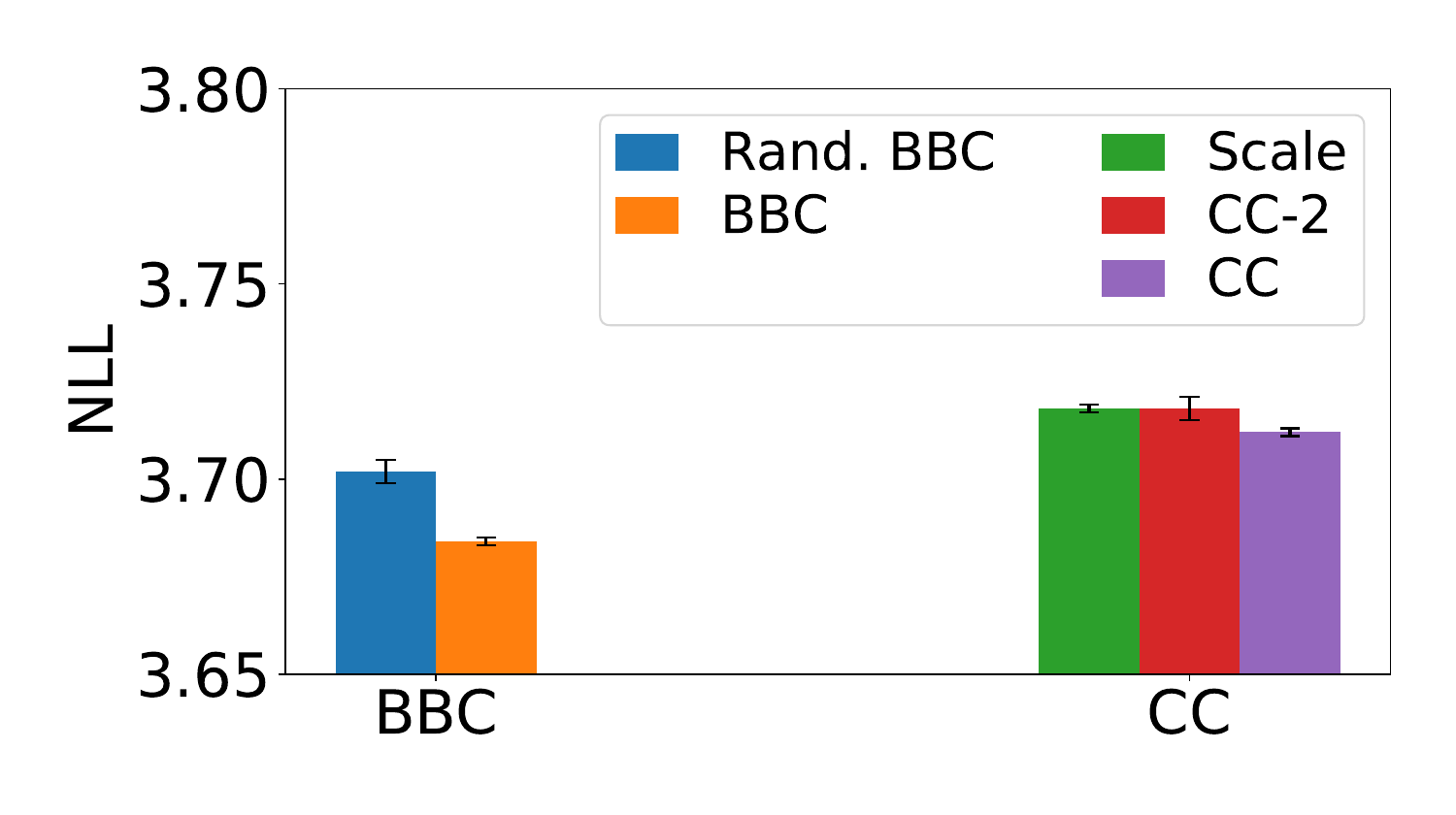}
        \caption{Ablations at 300M training tokens.}
        \label{fig:abl_300m}
    \end{subfigure}

    \caption{
        \textbf{Ablation results for BBC and CC.}
        At 100M tokens, BBC significantly outperforms sliding-window and random sparse attention, while CC significantly outperforms normalized roughness and cosine-similarity regularization. At 300M tokens, BBC significantly outperforms random BBC, while CC significantly outperforms both scale regularization and CC-2.
    }
    \label{fig:ablation_results}
\end{figure*}

\textbf{BBC is not merely sparse attention.}\ \ \  
BBC is a sparse attention pattern. 
To test whether its gains arise simply from reduced attention density, we compare BBC against same-density sliding-window and random sparse attention, applying all sparse patterns at layer 2. At 100M tokens, BBC significantly outperforms both
baselines, and neither of them improves over the vanilla baseline (Figure~\ref{fig:abl_100m}).

We further compare against \emph{Random BBC}, in which the carrier positions are selected at random while preserving the same per-token attention budget. At 300M tokens, BBC significantly outperforms Random BBC (Figure~\ref{fig:abl_300m}), indicating that the gain depends on structured carrier placement rather than the attention budget alone. Table~\ref{tab:bbc-ablations} in Appendix~\ref{app:bbc-ablations} reports the paired $t$-test $p$-values.

\textbf{CC requires both scale and roughness.}\ \ \ 
The CC objective has a natural tendency to reduce hidden-state scale (see Appendix~\ref{app:cc_scale_norm} for a formal discussion). To test whether its gains arise primarily from this scaling effect, we compare against a \emph{scale} regularizer,
$\mathcal{L}_{\rm scale} = \frac{1}{N}\sum_{i=1}^{N}\lVert h_i\rVert_2^2$,
and a \emph{normalized roughness} regularizer,
$\mathcal{L}_{\rm norm} = \frac{1}{N-1} \sum_{i=1}^{N-1} \frac{\lVert h_{i+1}-h_i\rVert_2^2}{\lVert h_{i+1}\rVert_2^2 + \lVert h_i\rVert_2^2 + \varepsilon}$. The normalized roughness term is complementary to scale in the CC objective.

CC is based on the path-graph Laplacian, whose Dirichlet energy is the weakest penalty in the finite-range pairwise family considered here (see Appendix~\ref{app:cc_k_connect}). We also include CC-2 $\mathcal{L}_{\rm CC\text{-}2}=\frac{1}{N-2}\sum_{i=1}^{N-2} \lVert 2h_{i+2}-h_{i+1}-h_i\rVert_2^2$, a two-predecessor Laplacian-residual penalty, in the ablation study. CC is also closely related to \emph{cosine distance} (see Appendix~\ref{app:cos}), which is included as well.

We tune the application layer and $\beta$ separately for each variant (Table~\ref{tab:scale_layer_sweep},~\ref{tab:norm_layer_sweep},~\ref{tab:cos_layer_sweep}, and~\ref{tab:cc2_layer_sweep} in Appendix~\ref{app:cc-ablations}). As shown in Figure~\ref{fig:ablation_results}, CC significantly outperforms all of these variants. Table~\ref{tab:cc-ablations} in Appendix~\ref{app:cc-ablations} reports the corresponding paired $t$-test $p$-values.

\textbf{20B training runs.}\ \ \ 
We next extend training to 20B tokens. We find that changing the learning-rate decay horizon also changes the optimal CC weight $\beta$. Following the same low-cost tuning strategy used in the 1B-token experiments, we select $\beta$ using a 1B-token run and find that its gain persists through 20B tokens (Figure~\ref{fig:teaser_scaling}).

During the 20B runs, however, BBC+CC temporarily underperforms BBC at intermediate checkpoints. We hypothesize that this is partly a numerical-stability issue: because $\beta \ll 1$, the effective contribution of the CC term can become numerically negligible as $\mathcal{L}_{\rm CC}$ decreases during training.

We therefore study two variants: \emph{random dropout}, which randomly omits the CC term on a fraction of optimization steps, and \emph{adaptive dropout}, which omits the CC term when its value falls below the $p$-th percentile of the preceding $h$-step history. The latter is designed to avoid repeatedly applying very small CC updates that may vanish after scaling by $\beta$.
With adaptive dropout, BBC+CC consistently outperforms BBC at every intermediate checkpoint (see Table~\ref{tab:20B-runs-loss-dedup-nocollision-10m} and Figure~\ref{fig:20B-runs-loss-dedup-no-collison} in Appendix~\ref{app:20b}).

\textbf{Representation diagnostics.}\ \ \ 
The preceding experiments establish that BBC and CC improve next-token
prediction. We next ask whether the learned representations change in the
directions suggested by Sections~\ref{sec:bbc} and~\ref{sec:cc}. 

\begin{figure}[h]
    \centering

    \begin{minipage}[t]{0.49\linewidth}
        \vspace{20pt}
        \centering

        \captionof{table}{
             Boundary-preference statistic $\Delta_b-\Delta_i$ at 100M
        training tokens under zero-vector and mean-replacement interventions.
        The last row reports the paired BBC--Vanilla $p$-value.
        }
        \label{tab:bbc_representation}

        \small
        \setlength{\tabcolsep}{3pt}
        \begin{tabular}{lcc}
        \toprule
        \textbf{Model} & \textbf{Zero replacement} & \textbf{Mean replacement} \\
        \midrule
        \textbf{Vanilla} & $-0.0025$ & $0.0017$ \\
        \textbf{BBC}     & $0.0052$  & $0.0065$ \\
        \midrule
        \textbf{$p$-value} & $3.46\times10^{-5}$ & $4.86\times10^{-4}$ \\
        \bottomrule
    \end{tabular}
    \end{minipage}
    \hfill
    \begin{minipage}[t]{0.49\linewidth}
        \vspace{0pt}
        \centering

        \includegraphics[width=\linewidth]{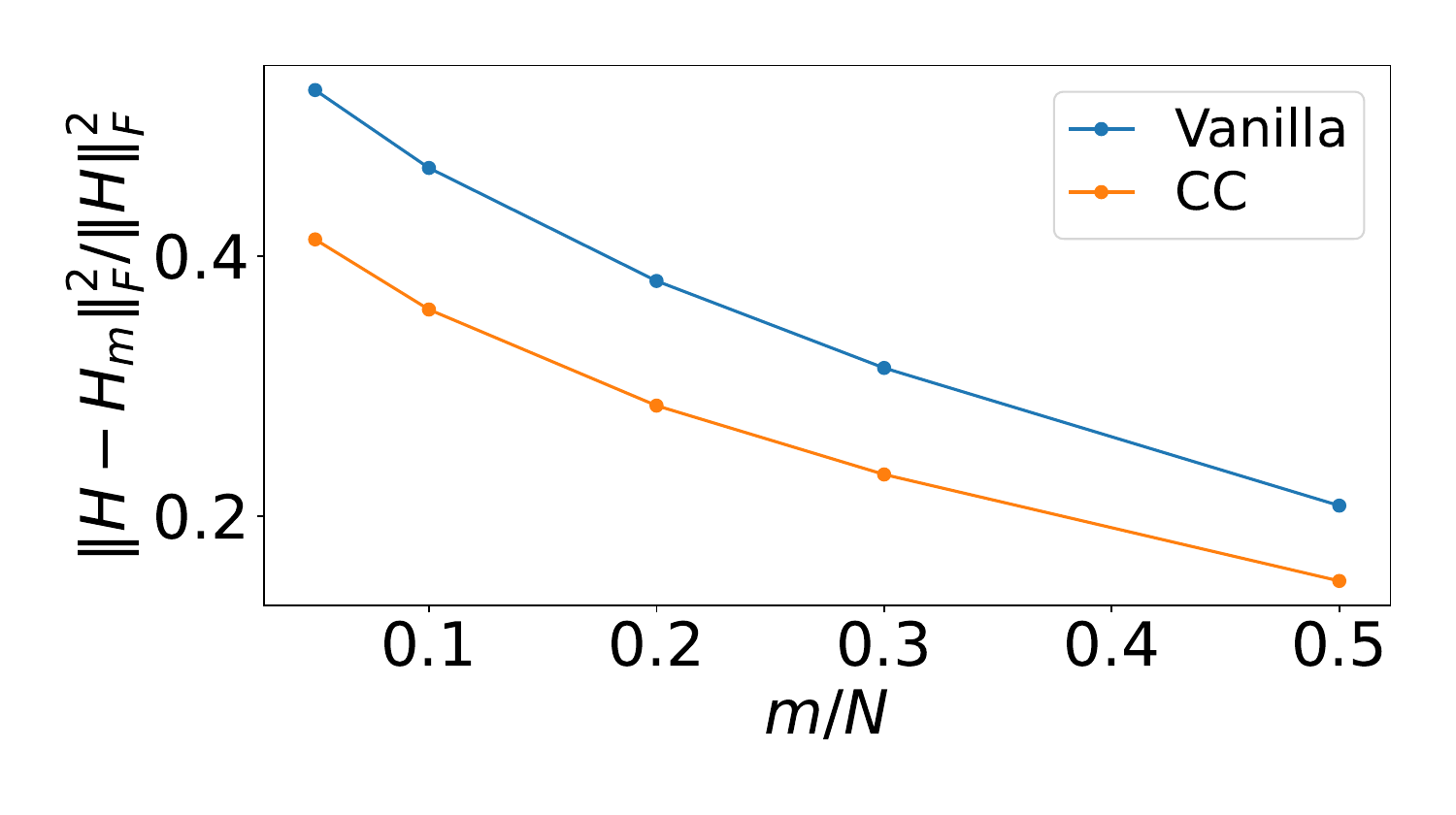}

        \captionof{figure}{
            \textbf{CC increases spectral compressibility.} Normalized spectral residual as a function of the retained
            low-frequency fraction $m/N$. 
        }
        \label{fig:cc_spectral_residual}
    \end{minipage}

\end{figure}

To test whether BBC increases the relative importance of boundary
representations, we either
replace the boundary representation and measure the resulting NLL increase,
$\Delta_b$, or replace a randomly selected non-boundary representation and
measure the corresponding increase, $\Delta_i$. A larger $\Delta_b - \Delta_i$ indicates that boundary representations are relatively more
consequential for prediction.
With both zero-vector replacement and replacement by the mean of the preceding
$K-1$ hidden states, BBC yields a significantly larger $\Delta_b - \Delta_i$ than Vanilla (Table~\ref{tab:bbc_representation}). In both experiments, we exclude
the first $W=4$ segments to control for the disproportionate influence of
initial-token positions (see Appendix~\ref{app:bbc_boundary_diagnostic} for full diagnostic).

Next we test whether reducing CC energy makes the
hidden-state trajectory better approximated by its low-frequency
path-graph modes.
For each hidden-state trajectory $H$, we project onto the first $m$
path-graph eigenmodes, $H_m = Q_m Q_m^\top H$,
and measure the normalized spectral residual
$\rho_m(H) = \frac{ \|H-H_m\|_F^2}{\|H\|_F^2}$.
This ratio is invariant to uniform rescaling and measures
the residual relative to total hidden-state energy. It includes the constant positional component; it does not isolate roughness of the centered trajectory.
CC reduces $\rho_m(H)$ at every measured spectral cutoff. Across the measured range, this corresponds to a
relative reduction of approximately $22\%$--$28\%$ (Figure~\ref{fig:cc_spectral_residual} and Table~\ref{tab:cc_spectral_residual} in Appendix~\ref{app:diag}).

\textbf{RoPE.}\ \ \ 
GPT-2 uses absolute positional embeddings, whereas most modern LLMs use
rotary positional embeddings (RoPE; \citealp{su2024roformer}). To test whether BBC and CC are robust
to the choice of positional encoding, we additionally train
\texttt{Pythia-410M} \citep{biderman2023pythia}, a GPT-2-like model using RoPE. We reuse the
\texttt{gpt2-medium} hyperparameters without retuning ($\eta=6\times10^{-4}$, $K=16$, and $\beta=0.00125$), and train for 1B tokens. BBC improves evaluation
NLL by 0.58\%, while BBC+CC improves it by 0.83\%, suggesting that the gains
are not specific to absolute positional embeddings (Table~\ref{tab:rope} in Appendix~\ref{app:rope-exp}).
We also sweep \texttt{rotary\_pct}, which controls the fraction of head
dimensions using rotary embeddings, over
$\{0, 0.25, 0.5, 0.75, 1.0\}$. The default value, $0.25$, also performs best
for \texttt{Pythia-410M} (Table~\ref{tab:pythia410m_100m} in
Appendix~\ref{app:rope-exp}).
RoPE modifies the attention mechanism; Appendix~\ref{app:rope} extends
Proposition~\ref{prop:boundary_attention} to rotary positional encoding.


\section{Related Work}
\label{sec:relatedwork}

\noindent \textbf{Representation learning for LLMs.}\ \ \ 
Representation learning has been central to self-supervised learning
in vision, where joint-embedding predictive architectures (JEPA)
\citep{assran2023self} learn semantic representations through
embedding-space prediction.
Recent work brings related ideas to language:
LLM-JEPA \citep{huang2025llmjepa} incorporates JEPA-style objectives
into language-model training, while Semantic Tube Prediction (STP)
\citep{huang2026stp} introduces a geometric prior on hidden-state
trajectories.
Large Concept Models \citep{barrault2024large} perform
autoregressive prediction in a sentence-level embedding space.
Our work shares the general motivation and introduces principled structure--residual inductive biases.

\noindent \textbf{Structured communication.}\ \ \ 
Fixed sparse attention \citep{child2019sparse} combines local attention with access to positions near previous block boundaries.
BBC can be viewed as a single-carrier ($c=1$) specialization---a setting
that \citet{child2019sparse} found overly restrictive---but we show that
applying it at the appropriate depth improves predictive
performance.
Block Transformer \citep{ho2024block} places
global block processing below local token decoding. Activation Beacon
\citep{zhang2025beacon} compresses context into inserted beacon tokens'
key--value states. H-Net \citep{hwang2026dynamic} learns boundaries, processes selected
states at reduced resolution, and upsamples for decoding. 
Local, random, and global connections are standard ingredients of
sparse attention designs such as Big Bird \citep{zaheer2020bigbird} and Landmark Attention~\citep{mohtashami2023landmark}. 

\noindent \textbf{Smoothness and spectral structure.}\ \ \ 
AWD-LSTM \citep{merity2018regularizing} penalizes successive recurrent-state
changes through temporal activation regularization.
\citet{li2026dispersion} encourage angular dispersion across batchwise hidden
states. \citet{shuman2013graphs} express graph smoothness through Laplacian
frequencies. CC regularizes adjacent Transformer hidden states 
and bounds low-frequency reconstruction error and finite-precision residual
code length, with ablations separating scale from positional roughness.

\noindent \textbf{Two-part coding.}\ \ \ 
CRATE \citep{yu2023whitebox} alternates token compression and
feature sparsification; ToST \citep{wu2025token} derives attention
by unrolling variational coding-rate reduction. \citet{deletang2024compression} turn predictive probabilities into lossless
data codes. 
Minimum description length \citep{rissanen1978modeling}
charges for the model and data encoded under it, while algorithmic statistics
\citep{gacs2001algorithmic} formalizes structure and realization through
two-part codes. We adopt this two-part accounting for BBC's segment realizations
and CC's spectral representations, retaining both structure and residual costs without
directly optimizing Kolmogorov complexity.

\noindent \textbf{Depth specialization.}\ \ \ 
\citet{skean2025layer} evaluate layerwise geometry and information content for
embedding quality. \citet{lad2025remarkable} probe computational sensitivity
by deleting and swapping layers. Our training-time layer sweeps identify
early BBC and late CC as favorable placements in two GPT-2 model sizes;
these preferences are observed empirically, not prescribed by the methods.

\section{Conclusion}
Motivated by the two-part coding view of Kolmogorov complexity, we introduced BBC and CC as complementary structure--residual inductive biases for language-model pretraining and derived explicit bounds characterizing both mechanisms. BBC compresses cross-segment communication through boundary carriers, while CC encourages compact spectral structure in hidden-state trajectories. We further give a qualitative connection between the two and find that BBC favors early layers, whereas CC favors later layers. Their gains persist across model sizes and training horizons up to 20B tokens, demonstrating that explicit representation biases can improve next-token pretraining.



\section*{Acknowledgments}

We thank Alexander Shen and Alexey Milovanov for early discussions on the two-part coding view of Kolmogorov complexity.

We gratefully acknowledge Polish highperformance computing infrastructure PLGrid
(HPC Center: ACK Cyfronet AGH) for providing computer facilities and support within
computational grant no. PLG/2025/018744.

\bibliography{iclr2027_conference}
\bibliographystyle{iclr2027_conference}

\clearpage
\appendix
\section{Additional Experimental Results}

\subsection{BBC and CC are depth-dependent}
\label{app:bbc-cc-layers}

\begin{table}[h]
\caption{Applying BBC at lower layers yields better NLL than at higher layers.}
\label{bbc-layers}

\begin{subtable}{\linewidth}
\centering
\label{bbc-layers-gpt-small}
\begin{tabular}{cccc}
\toprule
\small \textbf{Layer} & \small \textbf{$\pmb{K}$ = 4} & \small \textbf{$\pmb{K}$ = 8} & \small \textbf{$\pmb{K}$ = 16} \\
\midrule
\small 0 & \small 4.118$\pm$0.005 & \small 4.110$\pm$0.008 & \small 4.107$\pm$0.002 \\
\small 2 & \bf{\small 4.116$\pm$0.002} & \bf{\small 4.099$\pm$0.004} & \bf{\small 4.102$\pm$0.001} \\
\small 4 & \small 4.122$\pm$0.005 & \small 4.112$\pm$0.002  & \small 4.109$\pm$0.005 \\
\small 6 & \small 4.131$\pm$0.002 & \small 4.128$\pm$0.002  & \small 4.128$\pm$0.003 \\
\small 8 & \small 4.148$\pm$0.004 & \small 4.148$\pm$0.002 & \small 4.155$\pm$0.005 \\
\small 10 & \small 4.159$\pm$0.005 & \small 4.171$\pm$0.002 & \small 4.183$\pm$0.002 \\
\bottomrule
\end{tabular}
\vspace{0.25em}
\caption{\texttt{gpt2-small} at 100M tokens}
\end{subtable}

\vspace{1em}

\begin{subtable}{\linewidth}
\centering
\label{bbc-layers-gpt-medium}
\begin{tabular}{cccc}
\toprule
\small \textbf{Layer} & \small \textbf{$\pmb{K}$ = 4} & \small \textbf{$\pmb{K}$ = 8} & \small \textbf{$\pmb{K}$ = 16} \\
\midrule
\small 0 & \small 4.086$\pm$0.004 & \small 4.088$\pm$0.008 & \small 4.091$\pm$0.008 \\
\small 2 & \small 4.095$\pm$0.004 & \bf{\small 4.069$\pm$0.007} & \bf{\small 4.066$\pm$0.002} \\
\small 4 & \bf{\small 4.085$\pm$0.005} & \small 4.072$\pm$0.004 & \small 4.067$\pm$0.003 \\
\small 6 & \small 4.089$\pm$0.003 & \small 4.073$\pm$0.002 & \small 4.071$\pm$0.001 \\
\small 8 & \small 4.088$\pm$0.002 & \small 4.082$\pm$0.008 & \small 4.075$\pm$0.005 \\
\small 10 & \small 4.092$\pm$0.009 & \small 4.090$\pm$0.005 & \small 4.084$\pm$0.005 \\
\small 12 & \small 4.094$\pm$0.003 & \small 4.096$\pm$0.003 & \small 4.090$\pm$0.006 \\
\small 14 & \small 4.103$\pm$0.003 & \small 4.102$\pm$0.004 & \small 4.101$\pm$0.004 \\
\small 16 & \small 4.105$\pm$0.006 & \small 4.113$\pm$0.005 & \small 4.118$\pm$0.004 \\
\small 18 & \small 4.115$\pm$0.006 & \small 4.120$\pm$0.003 & \small 4.127$\pm$0.005 \\
\small 20 & \small 4.112$\pm$0.005 & \small 4.130$\pm$0.004 & \small 4.144$\pm$0.002 \\
\small 22 & \small 4.108$\pm$0.003 & \small 4.144$\pm$0.004 & \small 4.163$\pm$0.001 \\
\bottomrule
\end{tabular}
\vspace{0.25em}
\caption{\texttt{gpt2-medium} at 100M tokens}
\end{subtable}
\end{table}

\clearpage
Unlike BBC, which is applied within a Transformer layer, CC regularizes the hidden state between layers. To avoid introducing a layer index of $-1$, we index CC by the layer whose input it regularizes. Under this convention, the output of the final layer can be viewed as the input to a phantom layer $N+1$. We therefore sweep layers ${0,2,\dots,12}$ for \texttt{gpt2-small} and ${0,2,\dots,24}$ for \texttt{gpt2-medium} in Table~\ref{cc-layers}.

\begin{table}[h]
\caption{Applying $\mathcal{L}_{\rm CC}$ at higher layers yields better NLL than at lower layers.}
\label{cc-layers}

\begin{subtable}{\linewidth}
\centering
\label{cc-layers-gpt-small}
\begin{tabular}{cccc}
\toprule
\small \textbf{Layer} & \small \textbf{$\pmb{\beta}$ = 0.005} & \small \textbf{$\pmb{\beta}$ = 0.0025} & \small \textbf{$\pmb{\beta}$ = 0.00125} \\
\midrule
\small 12 & \small 4.139$\pm$0.003 & \small 4.146$\pm$0.002 & \small 4.144$\pm$0.007 \\
\small 10 & \bf{\small 4.113$\pm$0.002} & \bf{\small 4.106$\pm$0.002} & \bf{\small 4.112$\pm$0.002} \\
\small 8 & \small 4.115$\pm$0.003 & \small 4.113$\pm$0.005 & \small 4.117$\pm$0.001 \\
\small 6 & \small 4.121$\pm$0.004 & \small 4.122$\pm$0.003 & \small 4.122$\pm$0.000 \\
\small 4 & \small 4.129$\pm$0.004 & \small 4.130$\pm$0.003 & \small 4.141$\pm$0.003 \\
\small 2 & \small 4.137$\pm$0.001 & \small 4.142$\pm$0.003 & \small 4.146$\pm$0.002 \\
\small 0 & \small 4.149$\pm$0.001 & \small 4.143$\pm$0.002 & \small 4.143$\pm$0.001 \\
\bottomrule
\end{tabular}
\vspace{0.25em}
\caption{\texttt{gpt2-small} at 100M tokens}
\end{subtable}

\vspace{1em}

\begin{subtable}{\linewidth}
\centering
\label{cc-layers-gpt-medium}
\begin{tabular}{cccc}
\toprule
\small \textbf{Layer} & \small \textbf{$\pmb{\beta}$ = 0.005} & \small \textbf{$\pmb{\beta}$ = 0.0025} & \small \textbf{$\pmb{\beta}$ = 0.00125} \\
\midrule
\small 24 & \small 4.104$\pm$0.008 & \small 4.107$\pm$0.001 & \small 4.105$\pm$0.002 \\
\small 22 & \small 4.049$\pm$0.000 & \small 4.044$\pm$0.004 & \bf{\small 4.040$\pm$0.002} \\
\small 20 & \small 4.049$\pm$0.003 & \small 4.045$\pm$0.001 & \small 4.045$\pm$0.001 \\
\small 18 & \bf{\small 4.046$\pm$0.000} & \bf{\small 4.042$\pm$0.001} & \small 4.052$\pm$0.006 \\
\small 16 & \small 4.053$\pm$0.006 & \small 4.049$\pm$0.002 & \small 4.063$\pm$0.006 \\
\small 14 & \small 4.059$\pm$0.002 & \small 4.060$\pm$0.002 & \small 4.069$\pm$0.006 \\
\small 12 & \small 4.058$\pm$0.002 & \small 4.063$\pm$0.002 & \small 4.078$\pm$0.005 \\
\small 10 & \small 4.068$\pm$0.004 & \small 4.071$\pm$0.003 & \small 4.093$\pm$0.008 \\
\small 8 & \small 4.072$\pm$0.001 & \small 4.087$\pm$0.003 & \small 4.104$\pm$0.005 \\
\small 6 & \small 4.083$\pm$0.002 & \small 4.099$\pm$0.002 & \small 4.116$\pm$0.002 \\
\small 4 & \small 4.093$\pm$0.004 & \small 4.099$\pm$0.002 & \small 4.113$\pm$0.001 \\
\small 2 & \small 4.094$\pm$0.003 & \small 4.103$\pm$0.001 & \small 4.109$\pm$0.001 \\
\small 0 & \small 4.103$\pm$0.002 & \small 4.102$\pm$0.003 & \small 4.104$\pm$0.003 \\
\bottomrule
\end{tabular}
\vspace{0.25em}
\caption{\texttt{gpt2-medium} at 100M tokens}
\end{subtable}
\end{table}

\clearpage
\subsection{Optimal $K$ and $\beta$}
\label{app:convex}

\begin{table}[h]
\caption{BBC segment-length sweep at layer 2. Results are mean NLL $\pm$ standard deviation over three seeds.}
\label{tab:bbc_k_sweep}

\begin{subtable}{\linewidth}
\centering
\label{tab:bbc_k_sweep_small}
\begin{tabular}{ccccc}
\toprule
\small \textbf{Layer}
& \small \textbf{$K=32$}
& \small \textbf{$K=64$}
& \small \textbf{$K=128$}
& \small \textbf{$K=256$} \\
\midrule
\small 2
& \small 4.094$\pm$0.004
& \small 4.106$\pm$0.003
& \small 4.113$\pm$0.004
& \small 4.125$\pm$0.005 \\
\bottomrule
\end{tabular}
\vspace{0.25em}
\caption{\texttt{gpt2-small}}
\end{subtable}

\vspace{1em}

\begin{subtable}{\linewidth}
\centering
\label{tab:bbc_k_sweep_medium}
\begin{tabular}{ccccc}
\toprule
\small \textbf{Layer}
& \small \textbf{$K=32$}
& \small \textbf{$K=64$}
& \small \textbf{$K=128$}
& \small \textbf{$K=256$} \\
\midrule
\small 2
& \small 4.067$\pm$0.008
& \small 4.078$\pm$0.006
& \small 4.084$\pm$0.006
& \small 4.103$\pm$0.011 \\
\bottomrule
\end{tabular}
\vspace{0.25em}
\caption{\texttt{gpt2-medium}}
\end{subtable}
\end{table}

\begin{table}[h]
\caption{CC regularization-strength sweep. Results are mean NLL $\pm$ standard deviation over three seeds.}
\label{tab:cc_beta_sweep}

\begin{subtable}{\linewidth}
\centering
\label{tab:cc_beta_sweep_small}
\begin{tabular}{cccccc}
\toprule
\small \textbf{Layer}
& \small \textbf{$\beta=0.04$}
& \small \textbf{$\beta=0.02$}
& \small \textbf{$\beta=0.01$}
& \small \textbf{$\beta=0.000625$}
& \small \textbf{$\beta=0.0003125$} \\
\midrule
\small 10
& \small 4.142$\pm$0.007
& \small 4.114$\pm$0.001
& \small 4.112$\pm$0.003
& \small 4.115$\pm$0.004
& \small 4.121$\pm$0.003 \\
\bottomrule
\end{tabular}
\vspace{0.25em}
\caption{\texttt{gpt2-small}}
\end{subtable}

\vspace{1em}

\begin{subtable}{\linewidth}
\centering
\label{tab:cc_beta_sweep_medium}
\begin{tabular}{cccccc}
\toprule
\small \textbf{Layer}
& \small \textbf{$\beta=0.04$}
& \small \textbf{$\beta=0.02$}
& \small \textbf{$\beta=0.01$}
& \small \textbf{$\beta=0.000625$}
& \small \textbf{$\beta=0.0003125$} \\
\midrule
\small 22
& \small 4.087$\pm$0.009
& \small 4.060$\pm$0.008
& \small 4.045$\pm$0.001
& \small 4.047$\pm$0.000
& \small 4.056$\pm$0.003 \\
\bottomrule
\end{tabular}
\vspace{0.25em}
\caption{\texttt{gpt2-medium}}
\end{subtable}
\end{table}

\clearpage
\subsection{300M Significance Tests and 1B Persistence}
\label{app:300m-1b}

\begin{table}[h]
\caption{Results on \texttt{gpt2-small}. At 300M training tokens, CC, BBC, and BBC+CC all significantly outperform the vanilla baseline. The gains from BBC and BBC+CC persist through 1B training tokens.}
\label{small-300m-1b}

\begin{subtable}{\linewidth}
\centering
\label{tab:small-300m}
\begin{tabular}{c|cc|ccc}
\toprule
\small \textbf{Config} & \small \textbf{NLL} & \small \textbf{Improve (\%)} & \small \textbf{vanilla} & \small \textbf{CC} & \small \textbf{BBC} \\
\midrule
\small \textbf{vanilla} & \small 3.722$\pm$0.002 & \\
\small \textbf{CC} & \small 3.712$\pm$0.001 & \small 0.27 & \small 0.00228 \\
\small \textbf{BBC} & \small 3.685$\pm$0.002 & \small 0.99 & \small 0.000925 & \small 0.00109 \\
\small \textbf{BBC+CC} & \small 3.673$\pm$0.004 & \small 1.30 & \small 0.00231 & \small 0.00248 & \small 0.0364 \\
\bottomrule
\end{tabular}
\vspace{0.25em}
\caption{Results on \texttt{gpt2-small} at 300M tokens. The right-hand side of the table reports the $p$-values from paired $t$-tests.}
\end{subtable}

\vspace{1em}

\begin{subtable}{\linewidth}
\centering
\label{tab:small-1b}
\begin{tabular}{c|ccc}
\toprule
\small \textbf{Config} & \small \textbf{300M} & \small \textbf{600M} & \small \textbf{1B} \\
\midrule
\small \textbf{BBC} & \small 0.99$\pm$0.07 & \small 1.01$\pm$0.03 & \small 1.04$\pm$0.06 \\
\small \textbf{BBC+CC} & \small 1.30$\pm$0.15 & \small 1.24$\pm$0.13 & \small 1.36$\pm$0.10 \\
\bottomrule
\end{tabular}
\vspace{0.25em}
\caption{Results on \texttt{gpt2-small} through 1B tokens, reported as improvement (\%) over vanilla.}
\end{subtable}
\end{table}

\begin{table}[h]
\caption{Results on \texttt{gpt2-medium}. At 300M training tokens, CC, BBC, and BBC+CC all significantly outperform the vanilla baseline. The gains from BBC and BBC+CC persist through 1B training tokens.}
\label{medium-300m-1b}

\begin{subtable}{\linewidth}
\centering
\label{tab:medium-300m}
\begin{tabular}{c|cc|ccc}
\toprule
\small \textbf{Config} & \small \textbf{NLL} & \small \textbf{Improve (\%)} & \small \textbf{vanilla} & \small \textbf{CC} & \small \textbf{BBC} \\
\midrule
\small \textbf{vanilla} & \small 3.615$\pm$0.002 & \\
\small \textbf{CC} & \small 3.600$\pm$0.003 & \small 0.42 & \small 0.00286 \\
\small \textbf{BBC} & \small 3.575$\pm$0.002 & \small 1.11 & \small 0.0000675 & \small 0.000383 \\
\small \textbf{BBC+CC} & \small 3.563$\pm$0.003 & \small 1.44 & \small 0.000621 & \small 0.000466 & \small 0.00793 \\
\bottomrule
\end{tabular}
\vspace{0.25em}
\caption{Results on \texttt{gpt2-medium} at 300M tokens. The right-hand side of the table reports the $p$-values from paired $t$-tests.}
\end{subtable}

\vspace{1em}

\begin{subtable}{\linewidth}
\centering
\label{tab:medium-1b}
\begin{tabular}{c|ccc}
\toprule
\small \textbf{Config} & \small \textbf{300M} & \small \textbf{600M} & \small \textbf{1B} \\
\midrule
\small \textbf{BBC} & \small 1.11$\pm$0.02 & \small 1.19$\pm$0.04 & \small 1.25$\pm$0.04 \\
\small \textbf{BBC+CC} & \small 1.44$\pm$0.09 & \small 1.40$\pm$0.03& \small 1.51$\pm$0.02 \\
\bottomrule
\end{tabular}
\vspace{0.25em}
\caption{Results on \texttt{gpt2-medium} through 1B tokens, reported as improvement (\%) over vanilla.}
\end{subtable}
\end{table}

\clearpage
\subsection{BBC Ablations}
\label{app:bbc-ablations}

\begin{table}[h]
\caption{Results on \texttt{gpt2-small}. At 100M training tokens, sliding-window and random sparse attention both underperform the vanilla baseline, whereas BBC significantly outperforms it. At 300M tokens, BBC also significantly outperforms a variant in which the carrier positions are selected at random.}
\label{tab:bbc-ablations}

\begin{subtable}{\linewidth}
\centering
\label{tab:bbc-slide-sparse}
\begin{tabular}{c|cc|ccc}
\toprule
\small \textbf{Config} & \small \textbf{NLL} & \small \textbf{Improve (\%)} & \small \textbf{vanilla} & \small \textbf{sliding} & \small \textbf{rand sparse} \\
\midrule
\small \textbf{vanilla} & \small 4.142$\pm$0.002 & \\
\small \textbf{sliding} & \small 4.143$\pm$0.001 & \small $-$0.02 & \small 0.243 \\
\small \textbf{rand sparse} & \small 4.150$\pm$0.002 & \small $-$0.19 & \small 0.0372 & \small 0.0170 \\
\small \textbf{BBC} & \small 4.099$\pm$0.004 & \small 1.00 & \small 0.000313 & \small 0.00104 & \small 0.00220 \\
\bottomrule
\end{tabular}
\vspace{0.25em}
\caption{BBC vs. sliding-window attention and random sparse attention at 100M tokens. The right-hand side of the table reports the $p$-values from paired $t$-tests.}
\end{subtable}

\vspace{1em}

\begin{subtable}{\linewidth}
\centering
\label{tab:bbc-rand}
\begin{tabular}{c|c|c}
\toprule
\small \textbf{Config} & \small \textbf{NLL} & \small \textbf{$p$-value} \\
\midrule
\small \textbf{BBC} & \small 3.684$\pm$0.001 & \\
\small \textbf{rand BBC} & \small 3.702$\pm$0.003 & \small 0.0112 \\
\bottomrule
\end{tabular}
\vspace{0.25em}
\caption{BBC vs. random BBC at 300M tokens. The right-hand side of the table reports the $p$-values from paired $t$-tests.}
\end{subtable}
\end{table}

\subsection{CC Ablations}
\label{app:cc-ablations}

We tune the application layer and regularization strength $\beta$ separately for each CC variant---$\mathcal{L}_{\rm scale}$, $\mathcal{L}_{\rm norm}$, $1-\cos\theta$ as defined in Appendix~\ref{app:cos}, and $\mathcal{L}_{\rm CC\text{-}2}$. Because these objectives operate at different numerical scales and can exhibit different depth preferences, each requires its own matched choice of $\beta$ and application layer.

\paragraph{Scale regularization.}
As shown in Table~\ref{tab:scale_layer_sweep}, the best configuration applies the regularizer at layer 10 with $\beta=0.0025$.

\begin{table}[h]
\centering
\caption{$\mathcal{L}_{\rm scale}$ regularization on \texttt{gpt2-small} at 100M training tokens. Results are mean NLL $\pm$ standard deviation over three seeds.}
\label{tab:scale_layer_sweep}
\small
\begin{tabular}{c|ccc}
\toprule
\textbf{Layer}
& $\boldsymbol{\beta=0.00125}$
& $\boldsymbol{\beta=0.0025}$
& $\boldsymbol{\beta=0.005}$ \\
\midrule
0  & $4.146 \pm 0.005$ & $4.143 \pm 0.000$ & $4.145 \pm 0.005$ \\
2  & $4.137 \pm 0.000$ & $4.144 \pm 0.004$ & $4.147 \pm 0.005$ \\
4  & $4.126 \pm 0.005$ & $4.132 \pm 0.006$ & $4.132 \pm 0.008$ \\
6  & $4.119 \pm 0.003$ & $4.124 \pm 0.008$ & $4.125 \pm 0.005$ \\
8  & $4.120 \pm 0.002$ & $4.119 \pm 0.003$ & $4.120 \pm 0.001$ \\
10 & $\bf{4.119 \pm 0.005}$ & $\bf{4.111 \pm 0.003}$ & $\bf{4.114 \pm 0.004}$ \\
12 & $4.146 \pm 0.006$ & $4.144 \pm 0.002$ & $4.144 \pm 0.002$ \\
\bottomrule
\end{tabular}
\end{table}

\clearpage
\paragraph{Normalized roughness regularization.}
As shown in Table~\ref{tab:norm_layer_sweep}, the best configuration applies the regularizer at layer 2 with $\beta=0.000625$.

\begin{table}[h]
\centering
\caption{$\mathcal{L}_{\rm norm}$ regularization on \texttt{gpt2-small} at 100M training tokens. Results are mean NLL $\pm$ standard deviation over three seeds.}
\label{tab:norm_layer_sweep}
\small
\resizebox{\linewidth}{!}{
\begin{tabular}{c|ccccccc}
\toprule
\textbf{Layer}
& $\boldsymbol{\beta=0.0003125}$
& $\boldsymbol{\beta=0.000625}$
& $\boldsymbol{\beta=0.00125}$
& $\boldsymbol{\beta=0.0025}$
& $\boldsymbol{\beta=0.005}$
& $\boldsymbol{\beta=0.01}$
& $\boldsymbol{\beta=0.02}$ \\
\midrule
0
& $4.151 \pm 0.005$
& $4.145 \pm 0.004$
& $\bf{4.143 \pm 0.007}$
& $4.144 \pm 0.002$
& $4.146 \pm 0.003$
& $4.142 \pm 0.002$
& $4.144 \pm 0.007$ \\

2
& $4.146 \pm 0.002$
& $\bf{4.141 \pm 0.001}$
& $4.149 \pm 0.003$
& $4.146 \pm 0.007$
& $4.144 \pm 0.002$
& $4.150 \pm 0.005$
& $4.160 \pm 0.006$ \\

4
& $4.148 \pm 0.003$
& $4.147 \pm 0.002$
& $4.148 \pm 0.002$
& $4.151 \pm 0.005$
& $4.148 \pm 0.004$
& $4.145 \pm 0.006$
& $4.150 \pm 0.003$ \\

6
& $4.147 \pm 0.003$
& $4.149 \pm 0.002$
& $4.147 \pm 0.002$
& $4.147 \pm 0.001$
& $4.152 \pm 0.003$
& $4.146 \pm 0.005$
& $4.150 \pm 0.002$ \\

8
& $4.146 \pm 0.006$
& $4.146 \pm 0.002$
& $4.146 \pm 0.005$
& $4.145 \pm 0.004$
& $4.149 \pm 0.001$
& $4.146 \pm 0.004$
& $4.147 \pm 0.004$ \\

10
& $4.146 \pm 0.002$
& $4.144 \pm 0.008$
& $4.146 \pm 0.001$
& $4.147 \pm 0.007$
& $4.147 \pm 0.004$
& $4.148 \pm 0.005$
& $4.143 \pm 0.002$ \\

12
& $\bf{4.144 \pm 0.008}$
& $4.144 \pm 0.004$
& $4.144 \pm 0.005$
& $\bf{4.143 \pm 0.002}$
& $\bf{4.142 \pm 0.001}$
& $\bf{4.142 \pm 0.001}$
& $\bf{4.143 \pm 0.004}$ \\
\bottomrule
\end{tabular}
}
\end{table}

\paragraph{Cosine distance regularization.}
As shown in Table~\ref{tab:cos_layer_sweep}, the best configuration applies the regularizer at layer 8 with $\beta=0.0025$.

\begin{table}[h]
\centering
\caption{$1-\cos\theta$ regularization on \texttt{gpt2-small} at 100M training tokens. Results are mean NLL $\pm$ standard deviation over three seeds.}
\label{tab:cos_layer_sweep}
\small
\begin{tabular}{c|ccc}
\toprule
\textbf{Layer}
& $\boldsymbol{\beta=0.00125}$
& $\boldsymbol{\beta=0.0025}$
& $\boldsymbol{\beta=0.005}$ \\
\midrule
0  & $4.145 \pm 0.001$ & $4.148 \pm 0.006$ & $4.144 \pm 0.004$ \\
2  & $4.147 \pm 0.006$ & $4.145 \pm 0.010$ & $4.145 \pm 0.002$ \\
4  & $4.144 \pm 0.001$ & $4.146 \pm 0.003$ & $4.146 \pm 0.006$ \\
6  & $4.144 \pm 0.002$ & $4.145 \pm 0.001$ & $4.153 \pm 0.002$ \\
8  & $4.146 \pm 0.003$ & $\bf{4.142 \pm 0.004}$ & $4.145 \pm 0.004$ \\
10 & $\bf{4.142 \pm 0.001}$ & $4.144 \pm 0.006$ & $4.146 \pm 0.007$ \\
12 & $4.146 \pm 0.004$ & $4.144 \pm 0.002$ & $\bf{4.143 \pm 0.002}$ \\
\bottomrule
\end{tabular}
\end{table}

\paragraph{CC-2 regularization.}
As shown in Table~\ref{tab:cc2_layer_sweep}, the best configuration applies the regularizer at layer 10 with $\beta=0.0025$.

\begin{table}[h]
\centering
\caption{$\mathcal{L}_{\rm CC\text{-}2}$ regularization on \texttt{gpt2-small} at 100M training tokens. Results are mean NLL $\pm$ standard deviation over three seeds.}
\label{tab:cc2_layer_sweep}
\small
\begin{tabular}{c|ccc}
\toprule
\textbf{Layer}
& $\boldsymbol{\beta=0.00125}$
& $\boldsymbol{\beta=0.0025}$
& $\boldsymbol{\beta=0.005}$ \\
\midrule
0  & $4.145 \pm 0.002$ & $4.145 \pm 0.003$ & $4.146 \pm 0.001$ \\
2  & $4.147 \pm 0.003$ & $4.156 \pm 0.007$ & $4.178 \pm 0.004$ \\
4  & $4.135 \pm 0.005$ & $4.148 \pm 0.002$ & $4.179 \pm 0.006$ \\
6  & $4.122 \pm 0.002$ & $4.131 \pm 0.007$ & $4.160 \pm 0.008$ \\
8  & $4.117 \pm 0.005$ & $4.121 \pm 0.003$ & $4.135 \pm 0.005$ \\
10 & $\bf{4.110 \pm 0.001}$ & $\bf{4.110 \pm 0.010}$ & $\bf{4.120 \pm 0.003}$ \\
12 & $4.148 \pm 0.003$ & $4.145 \pm 0.002$ & $4.144 \pm 0.007$ \\
\bottomrule
\end{tabular}
\end{table}

\clearpage
\paragraph{Significance tests.}
At 100M training tokens, CC significantly outperforms both cosine-distance and normalized-roughness regularization. For scale regularization and CC-2, the performance gap becomes statistically significant only after extending training to 300M tokens.

\begin{table}[h]
\caption{Results on \texttt{gpt2-small}. At 100M training tokens, CC significantly outperforms both normalized roughness and cosine similarity regularization. At 300M tokens, CC also significantly outperforms scale regularization and CC-2.}
\label{tab:cc-ablations}

\begin{subtable}{\linewidth}
\centering
\label{tab:cc-norm-cos}
\begin{tabular}{c|cc|ccc}
\toprule
\small \textbf{Config} & \small \textbf{NLL} & \small \textbf{Improve (\%)} & \small \textbf{vanilla} & \small \textbf{cosine} & \small \textbf{normalized} \\
\midrule
\small \textbf{vanilla} & \small 4.142$\pm$0.002 & \\
\small \textbf{cosine} & \small 4.142$\pm$0.004 & \small 0.01 & \small 0.466 \\
\small \textbf{normalized} & \small 4.141$\pm$0.001 & \small 0.02 & \small 0.288 & \small 0.412 \\
\small \textbf{CC} & \small 4.106$\pm$0.002 & \small 0.87 & \small 0.00127 & \small 0.00154 & \small 0.0000978 \\
\bottomrule
\end{tabular}
\vspace{0.25em}
\caption{CC vs. cosine similarity and normalized roughness regularization at 100M tokens. The right-hand side of the table reports the $p$-values from paired $t$-tests.}
\end{subtable}

\vspace{1em}

\begin{subtable}{\linewidth}
\centering
\label{tab:cc-scale-cc2}
\begin{tabular}{c|c|cc}
\toprule
\small \textbf{Config} & \small \textbf{NLL} & \small \textbf{scale} & \small \textbf{CC-2} \\
\midrule
\small \textbf{scale} & \small 3.718$\pm$0.001 & \\
\small \textbf{CC-2} & \small 3.718$\pm$0.003 & \small 0.477 \\
\small \textbf{CC} & \small 3.712$\pm$0.001 & \small 0.0113 & \small 0.0386 \\
\bottomrule
\end{tabular}
\vspace{0.25em}
\caption{CC vs. scale regularization and CC-2 at 300M tokens. The right-hand side of the table reports the $p$-values from paired $t$-tests.}
\end{subtable}
\end{table}

\clearpage
\subsection{20B Training Runs}
\label{app:20b}

When extending training to 20B tokens, we first retune the CC weight $\beta$
using a 1B-token sweep. Under the longer schedule, the slower learning-rate
decay allows CC to exert greater cumulative downward pressure on
$\mathcal{L}_{\rm CC}$, which can shift the optimal $\beta$. Using a
single seed (82), we find that $\beta=0.000625$ performs best at 1B tokens.
With this setting, BBC+CC eventually outperforms BBC after 10B tokens, but
underperforms it at intermediate checkpoints between 4B and 9B tokens.
Further decreasing $\beta$ does not improve the 1B-token result.

We hypothesize that this behavior is partly due to numerical stability:
as $\mathcal{L}_{\rm CC}$ decreases, multiplying it by $\beta\ll1$ can make
the corresponding optimization signal numerically negligible. We therefore
study two variants. \emph{Random dropout} skips the CC term independently with
dropout rate $d>0$. \emph{Adaptive dropout} instead skips the CC term whenever
$\mathcal{L}_{\rm CC}$ falls below the $p$-th percentile of its values over the
preceding $h$ optimization steps, reducing the frequency of very small CC
updates.

At 1B training tokens, we evaluate random dropout with $d=0.5$ and adaptive
dropout with $p=50$ and $h=32$ using three seeds
$\{82,23,37\}$ (Table~\ref{tab:20b-sweep}). Both variants substantially
reduce the across-seed variance relative to standard CC, and adaptive dropout
achieves the best mean performance of the three. We therefore use adaptive
dropout for a second 20B-token run, where BBC+CC Adapt outperforms BBC at every
intermediate checkpoint.

\begin{figure}[h]
    \centering
    \includegraphics[width=\linewidth]{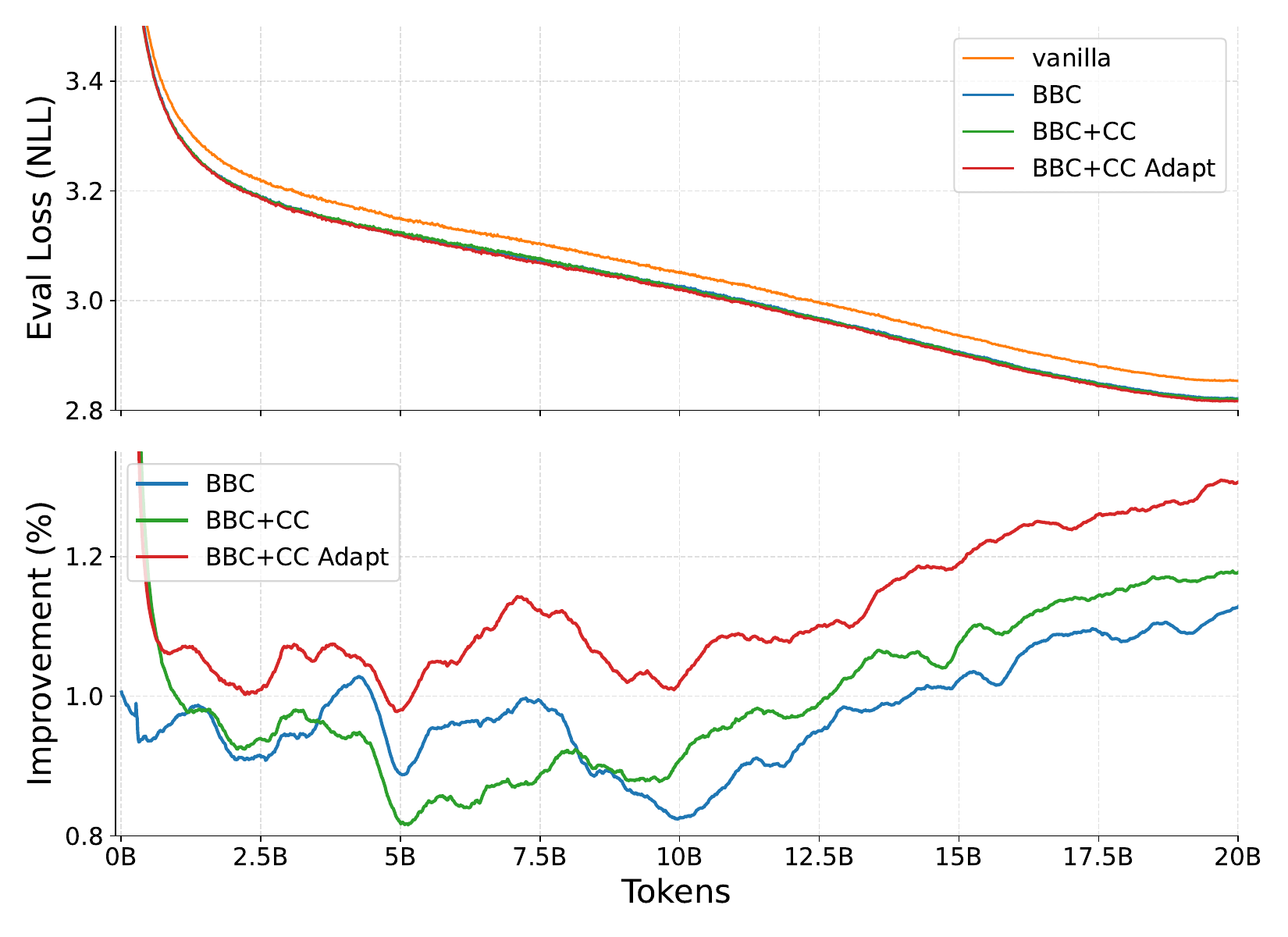}
    \caption{\textbf{Metrics for 20B runs.} Evaluation NLL loss and Improvement (\%) over vanilla during 20B \texttt{gpt2-medium} training runs. The evaluation is run on 1M-token holdout set every 10M training tokens.}
    \label{fig:20B-runs-loss-dedup-no-collison}
\end{figure}

We evaluate every 1B-token intermediate checkpoint on an enlarged 10M-token holdout set to reduce evaluation noise. This gives the high precision required for comparing CC variants (Table~\ref{tab:20b-sweep}) and offsets the absence of seed averaging in the single-seed 20B-token runs (Table~\ref{tab:20B-runs-loss-dedup-nocollision-10m}). In addition, we evaluate on the 1M-token holdout set every 10M training tokens, which produces Figure~\ref{fig:20B-runs-loss-dedup-no-collison}.

\begin{table}[h]
\centering
\caption{BBC+CC variants for \texttt{gpt2-medium} at 1B tokens with 20B learning-rate decay horizon and $\beta=0.000625$. For \label{tab:20b-sweep}
\emph{Random Dropout} the probability is 0.5. For \emph{Adaptive Dropout} the $p$-th percentile is 50 and history size is 32. The evaluation is run on 10M-token holdout set.}
\begin{tabular}{l|c}
\toprule
\small \textbf{BBC+CC Variant} & \small \textbf{NLL} \\
\midrule
\small Standard CC & \small 3.2757$\pm$0.0043 \\
\small CC Random Dropout $d=0.5$ & \small 3.2757$\pm$0.0008 \\
\small CC Adaptive Dropout $p=50,\ h=32$& \small 3.2753$\pm$0.0005 \\
\bottomrule
\end{tabular}
\end{table}

\begin{table}[h]
\caption{Results on \texttt{gpt2-medium} at 20B tokens. BBC and both BBC+CC variants outperform the vanilla baseline at every checkpoint (single seed, so no significance test). BBC+CC lags behind BBC at some initial checkpoints while BBC+CC Adapt consistently outperforms it. Each checkpoint is evaluated on 10M-token holdout set.}
\label{tab:20B-runs-loss-dedup-nocollision-10m}
\begin{center}
\begin{tabular}{c|ccccccccc}
\toprule
&& \multicolumn{4}{c}{\small \textbf{NLL}} && \multicolumn{3}{c}{\small \textbf{Improve (\%)}} \\[0.7ex]
\small \textbf{Tokens} && \small vanilla & \small BBC & \small BBC+CC & \small BBC+CC Adapt && \small BBC & \small BBC+CC & \small BBC+CC Adapt \\
\midrule
\small 1B  && \small 3.3113 & \small 3.2772 & \small 3.2765 & \small 3.2748 && \small 1.03 & \small 1.05 & \small 1.10 \\
\small 2B  && \small 3.2107 & \small 3.1819 & \small 3.1807 & \small 3.1793 && \small 0.90 & \small 0.94 & \small 0.98 \\
\small 3B  && \small 3.1704 & \small 3.1397 & \small 3.1387 & \small 3.1363 && \small 0.97 & \small 1.00 & \small 1.08 \\
\small 4B  && \small 3.1426 & \small 3.1089 & \small 3.1125 & \small 3.1085 && \small 1.08 & \small 0.96 & \small 1.09 \\
\small 5B  && \small 3.1184 & \small 3.0897 & \small 3.0937 & \small 3.0871 && \small 0.92 & \small 0.79 & \small 1.00 \\
\small 6B  && \small 3.0980 & \small 3.0671 & \small 3.0720 & \small 3.0649 && \small 1.00 & \small 0.84 & \small 1.07 \\
\small 7B  && \small 3.0801 & \small 3.0494 & \small 3.0533 & \small 3.0461 && \small 1.00 & \small 0.87 & \small 1.10 \\
\small 8B  && \small 3.0609 & \small 3.0304 & \small 3.0315 & \small 3.0268 && \small 0.99 & \small 0.96 & \small 1.11 \\
\small 9B  && \small 3.0376 & \small 3.0116 & \small 3.0118 & \small 3.0072 && \small 0.86 & \small 0.85 & \small 1.00 \\
\small 10B && \small 3.0185 & \small 2.9922 & \small 2.9916 & \small 2.9872 && \small 0.87 & \small 0.89 & \small 1.04 \\
\small 11B && \small 2.9983 & \small 2.9700 & \small 2.9700 & \small 2.9660 && \small 0.95 & \small 0.94 & \small 1.08 \\
\small 12B && \small 2.9757 & \small 2.9475 & \small 2.9467 & \small 2.9438 && \small 0.95 & \small 0.97 & \small 1.07 \\
\small 13B && \small 2.9526 & \small 2.9221 & \small 2.9214 & \small 2.9194 && \small 1.03 & \small 1.06 & \small 1.12 \\
\small 14B && \small 2.9290 & \small 2.8984 & \small 2.8971 & \small 2.8944 && \small 1.04 & \small 1.09 & \small 1.18 \\
\small 15B && \small 2.9036 & \small 2.8733 & \small 2.8722 & \small 2.8703 && \small 1.04 & \small 1.08 & \small 1.15 \\
\small 16B && \small 2.8784 & \small 2.8473 & \small 2.8462 & \small 2.8433 && \small 1.08 & \small 1.12 & \small 1.22 \\
\small 17B && \small 2.8569 & \small 2.8249 & \small 2.8245 & \small 2.8223 && \small 1.12 & \small 1.14 & \small 1.21 \\
\small 18B && \small 2.8379 & \small 2.8063 & \small 2.8048 & \small 2.8028 && \small 1.11 & \small 1.17 & \small 1.24 \\
\small 19B && \small 2.8244 & \small 2.7930 & \small 2.7908 & \small 2.7886 && \small 1.11 & \small 1.19 & \small 1.27 \\
\small 20B && \small 2.8193 & \small 2.7876 & \small 2.7858 & \small 2.7837 && \small 1.12 & \small 1.19 & \small 1.26 \\
\bottomrule
\end{tabular}
\end{center}
\end{table}

\clearpage
\subsection{Boundary Intervention Diagnostic}
\label{app:bbc_boundary_diagnostic}

To test whether BBC increases the relative importance of boundary
representations, we evaluate frozen 100M-token Vanilla and BBC checkpoints
using two matched activation interventions. For each completed segment, we
either replace the boundary representation by a zero vector and measure the
resulting increase in NLL, denoted $\Delta_b$, or replace a randomly selected
non-boundary representation in the same segment and measure the corresponding
increase, denoted $\Delta_i$. We use
\begin{equation}
    G
    =
    \Delta_b-\Delta_i
\end{equation}
as the boundary-preference statistic. A positive $G$ indicates that ablating
the boundary is more damaging than ablating a matched interior position. More
importantly,
\begin{equation}
    G_{\mathrm{BBC}} > G_{\mathrm{Vanilla}}
\end{equation}
indicates that BBC increases the relative predictive importance of boundary
representations.

\begin{figure}[h]
    \centering
    \includegraphics[width=0.8\linewidth]{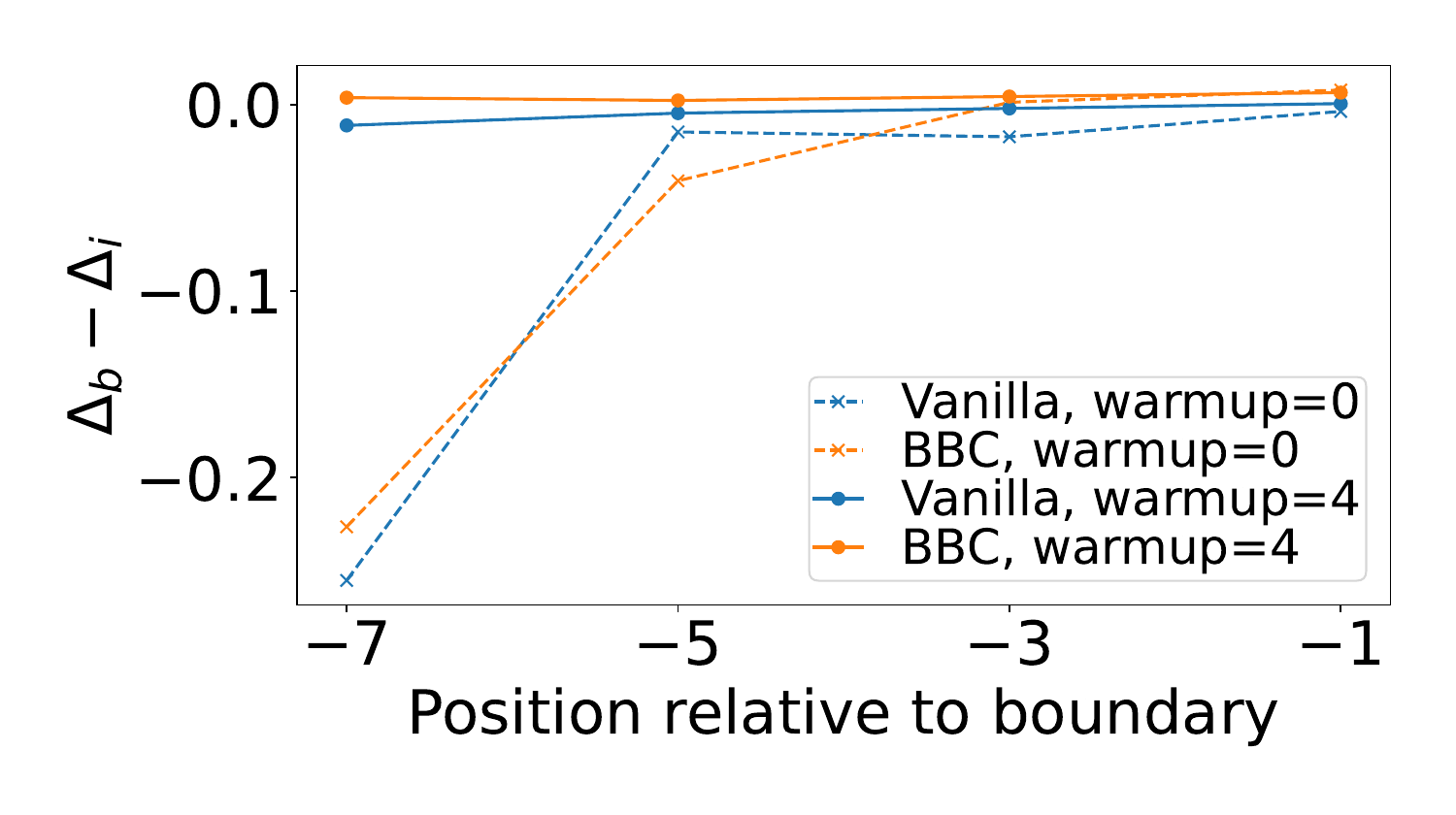}
    \caption{
        \textbf{Effect of the ablated interior position on the boundary-preference
        statistic $\Delta_b-\Delta_i$.} Without excluding the first segments
        ($W=0$), ablating the earliest position produces a large positional
        artifact. Excluding the first four segments ($W=4$) largely removes
        this dependence.
    }
    \label{fig:zero_position}
\end{figure}

\paragraph{Controlling for early-position effects.}
A direct intervention over the entire sequence is confounded by the unusually
large influence of the first few tokens. Figure~\ref{fig:zero_position} shows
the boundary-preference statistic as the ablated interior position is moved
relative to the segment boundary. With $K=8$, position $-7$ corresponds to
the first token of a segment. Without excluding the initial segments, ablating
this position produces a disproportionately large change in NLL, dominating
the boundary-versus-interior comparison.

We therefore exclude the first $W=4$ segments from both the intervention
statistics and the corresponding NLL evaluation. After this warmup exclusion,
the dependence on the particular non-boundary position becomes negligible
(Figure~\ref{fig:zero_position}), allowing the diagnostic to isolate the
relative effect of boundary placement rather than the special role of the
earliest tokens.

\paragraph{Zero-vector intervention.}
With $W=4$, we obtain
\begin{center}
\begin{tabular}{lccc}
\toprule
Model
& $\Delta_b$
& $\Delta_i$
& $\Delta_b-\Delta_i$ \\
\midrule
Vanilla
& $1.2293 \pm 0.1566$
& $1.2318 \pm 0.1556$
& $-0.0025 \pm 0.0618$ \\
BBC
& $0.8553 \pm 0.1028$
& $0.8500 \pm 0.1026$
& $0.0052 \pm 0.0589$ \\
\bottomrule
\end{tabular}
\end{center}

Vanilla exhibits essentially no positive boundary preference:
$\Delta_b-\Delta_i=-0.0025$. In contrast, BBC yields a positive
boundary-preference gap of $0.0052$. More importantly, the paired
BBC--Vanilla contrast is statistically significant,
\begin{equation}
    t(1023)=4.160,
    \qquad
    p=3.46\times10^{-5}.
\end{equation}
Thus BBC significantly increases the relative sensitivity of the model to
boundary representations compared with matched interior representations.

\paragraph{Mean-replacement intervention.}
To verify that the result is not specific to zero-vector ablation, we repeat
the experiment using a less disruptive intervention: the selected
representation is replaced by the mean of the preceding $K-1$ hidden states.
We obtain
\begin{center}
\begin{tabular}{lccc}
\toprule
Model
& $\Delta_b$
& $\Delta_i$
& $\Delta_b-\Delta_i$ \\
\midrule
Vanilla
& $0.6267 \pm 0.0683$
& $0.6251 \pm 0.0670$
& $0.0017 \pm 0.0626$ \\
BBC
& $0.5612 \pm 0.0636$
& $0.5546 \pm 0.0616$
& $0.0065 \pm 0.0584$ \\
\bottomrule
\end{tabular}
\end{center}

The Vanilla gap is small relative to its per-example variability, whereas BBC
again exhibits a larger positive boundary preference. The paired contrast is
significant:
\begin{equation}
    t(1023)=3.499,
    \qquad
    p=4.86\times10^{-4}.
\end{equation}

Taken together, the two interventions give the same qualitative result:
standard next-token training shows little or no systematic preference for the
fixed segment boundaries used by BBC, whereas BBC significantly increases the
relative predictive importance of those boundary representations. This
provides direct representation-level evidence that the BBC routing constraint
causes useful information to become preferentially concentrated at the
designated carrier positions.

\subsection{Spectral diagnostics for CC}
\label{app:diag}

\begin{table}[h]
\centering
\caption{Normalized spectral residual
$\|H-H_m\|_F^2/\|H\|_F^2$ at 100M training tokens. Lower is better
approximated by the first $m$ positional modes.}
\label{tab:cc_spectral_residual}
\begin{tabular}{lccccc}
\toprule
\small \textbf{Model}
& \small $m/N=.05$
& \small $m/N=.10$
& \small $m/N=.20$
& \small $m/N=.30$
& \small $m/N=.50$ \\
\midrule
\small \textbf{Vanilla}
& \small 0.528
& \small 0.468
& \small 0.381
& \small 0.314
& \small 0.208 \\
\small \textbf{CC}
& \small 0.413
& \small 0.359
& \small 0.285
& \small 0.232
& \small 0.150 \\
\bottomrule
\end{tabular}
\end{table}

\subsection{RoPE}
\label{app:rope-exp}

\begin{table}[h]
\centering
\caption{Results on \texttt{Pythia-410M} after 1B training tokens using RoPE. Results are mean NLL $\pm$ standard deviation over three seeds. We reuse the \texttt{gpt2-medium} hyperparameters:
$\eta=6\times10^{-4}$, $K=16$, and $\beta=0.00125$.}
\label{tab:rope}
\begin{tabular}{lcc}
\toprule
\small \textbf{Model}
& \small \textbf{NLL}
& \small \textbf{Improve (\%)} \\
\midrule
\small \textbf{Vanilla}
& \small $3.1883 \pm 0.001$
& \small -- \\
\small \textbf{BBC}
& \small $3.1698 \pm 0.002$
& \small $0.58$ \\
\small \textbf{BBC+CC}
& \small $3.1620 \pm 0.001$
& \small $0.83$ \\
\bottomrule
\end{tabular}
\end{table}

\begin{table}[t]
\centering
\caption{Impact of RoPE's \texttt{rotary\_pct} (fraction of head dimensions with rotary embeddings), changed only at the BBC layer ($L=2$), on \texttt{Pythia-410M} performance after 100M tokens. Results are mean NLL $\pm$ standard deviation over three seeds; improvement is relative to the vanilla baseline.}
\begin{tabular}{lcc}
\toprule
\small \textbf{Model} & \small \textbf{NLL} & \small \textbf{Improve (\%)} \\
\midrule
\small \textbf{Vanilla}\ \ \ \ \ 
\texttt{rotary\_pct}=0.25          & \small $4.0844 \pm 0.006$ & \small -- \\
\small \textbf{BBC+CC} \small \texttt{rotary\_pct}=0.00     & \small $4.0460 \pm 0.006$ & \small 0.95 \\
\small \textbf{BBC+CC}  \texttt{rotary\_pct}=0.25     & \small $4.0066 \pm 0.001$ & \small 1.94 \\
\small \textbf{BBC+CC}  \texttt{rotary\_pct}=0.50     & \small $4.0066 \pm 0.004$ & \small 1.94 \\
\small \textbf{BBC+CC}  \texttt{rotary\_pct}=0.75     & \small $4.0074 \pm 0.001$ & \small 1.92 \\
\small \textbf{BBC+CC}  \texttt{rotary\_pct}=1.00     & \small $4.0077 \pm 0.003$ & \small 1.91 \\
\bottomrule
\end{tabular}
\label{tab:pythia410m_100m}
\end{table}

\clearpage
\section{Additional Connections Between BBC and CC}
\label{app:bbc_cc_connection}

We now consider the reverse direction. BBC restricts the
value subspace available for cross-segment communication.
Smoothness requires an additional condition on how attention
changes between neighboring queries.

\subsection{Cross-segment feature subspace}
\label{app:bbc_rank}

Consider one deterministic attention head. Let $B$ be the
union of boundary positions available to the queries,
assigning zero weights to inaccessible positions. Let
\[
A_B\in\mathbb{R}^{N_q\times|B|},
\qquad
V_B\in\mathbb{R}^{|B|\times d_v}.
\]
The cross-segment contribution is
\begin{equation}
    O_{\mathrm{cross}}=A_BV_B.
    \label{eq:bbc_cross_output}
\end{equation}
Rows of $A_B$ need not sum to one, because local-prefix
tokens also receive attention.

\begin{proposition}[Cross-segment subspace constraint]
\label{prop:bbc_rank}
\begin{equation}
    \operatorname{rank}(O_{\mathrm{cross}})
    \leq\min(N_q,|B|,d_v),
    \label{eq:bbc_rank_bound}
\end{equation}
and every output row belongs to
$\operatorname{span}\{v_b:b\in B\}$.
\end{proposition}

\begin{proof}
Each row of $A_BV_B$ is a linear combination of boundary
values, and the rank of a product cannot exceed the rank
of either factor.
\end{proof}

This constrains one head's cross-segment contribution,
rather than the complete hidden state. The carrier-count
bound need not tighten the value-dimension bound when
$|B|\geq d_v$. Nor does low rank imply smoothness:
alternating rows $u,-u,u,-u,\ldots$ have rank one and
adjacent energy $4(N_q-1)\|u\|_2^2$.

\subsection{Conditional smoothness}
\label{app:bbc_to_cc}

For the adjacent-difference operator $D$,
\begin{equation}
    \|DO_{\mathrm{cross}}\|_F^2
    \leq
    \|V_B\|_{\mathrm{op}}^2\|DA_B\|_F^2.
    \label{eq:bbc_attention_smoothness}
\end{equation}
Thus similar boundary-attention weights yield similar
cross-segment outputs.

For this comparison, restrict $B$ to the fixed, nonempty set of
boundaries preceding one segment. Let
$K_B\in\mathbb{R}^{|B|\times d_k}$ contain their keys as rows,
and restrict $V_B$ to the same positions. Assume boundary logits
$q_i^\top k_b/\sqrt{d_k}$. Define the boundary-normalized distribution
and total boundary attention mass:
\begin{equation}
    \overline a_i=
    \operatorname{softmax}
    \left(\frac{K_Bq_i}{\sqrt{d_k}}\right),
    \qquad
    \gamma_i=\sum_{b\in B}(A_B)_{ib}.
    \label{eq:bbc_boundary_attention}
\end{equation}
Then
\begin{equation}
    z_i=V_B^\top\overline a_i,
    \qquad
    o_i^{\mathrm{cross}}=\gamma_i z_i.
    \label{eq:bbc_boundary_mass}
\end{equation}
Here $0\leq\gamma_i\leq1$; the remaining attention mass
belongs to the local prefix.

Write $E_{\mathrm{intra}}$ for adjacent energy restricted
to this segment, and let $Q^{\mathrm{qry}}$ be the query
matrix. Set
\[
V_{\max}=\max_{b\in B}\|v_b\|_2,
\qquad
C_B=
\frac{\|V_B\|_{\mathrm{op}}\|K_B\|_{\mathrm{op}}}
     {2\sqrt{d_k}}.
\]

\begin{proposition}[Conditional BBC-to-CC smoothness]
\label{prop:bbc_smoothness}
For fixed boundary keys and values,
\begin{equation}
\begin{aligned}
E_{\mathrm{intra}}(O_{\mathrm{cross}})
\leq
\bigl(
V_{\max}\sqrt{E_{\mathrm{intra}}(\gamma)}
+
C_B\sqrt{E_{\mathrm{intra}}(Q^{\mathrm{qry}})}
\bigr)^2.
\end{aligned}
    \label{eq:bbc_to_cc_bound}
\end{equation}
\end{proposition}

\begin{proof}
The softmax Jacobian is
$J=\operatorname{diag}(p)-pp^\top$. It is symmetric and
has absolute row sums $2p_i(1-p_i)\leq1/2$, so its
Euclidean operator norm is at most $1/2$. Thus
\[
\|z_{i+1}-z_i\|_2
\leq C_B\|q_{i+1}-q_i\|_2.
\]
Using $o_i^{\mathrm{cross}}=\gamma_i z_i$,
$\|z_i\|_2\leq V_{\max}$, and $0\leq\gamma_i\leq1$ gives
\[
\|o_{i+1}^{\mathrm{cross}}-o_i^{\mathrm{cross}}\|_2
\leq
V_{\max}|\gamma_{i+1}-\gamma_i|
+
C_B\|q_{i+1}-q_i\|_2.
\]
The triangle inequality over consecutive differences
gives the result.
\end{proof}

Both terms matter. With one boundary value $u$, zero
local values, and identical zero logits, the cross-segment
output changes from $u/2$ to $u/3$ when the local prefix
grows from one token to two, even with identical queries.

BBC therefore transfers smoothness when both queries
and total boundary attention mass vary smoothly.
Transitions between segments, where the boundary set
changes, require a separate term.

\subsection{\texorpdfstring{Proposition~\ref{prop:boundary_attention} at trained BBC layers}{Proposition 4 at trained BBC layers}}
\label{app:prop4_check}
We test Proposition~\ref{prop:boundary_attention} at the BBC layer (layer~2) of two models trained for 1B tokens: \texttt{gpt2-medium} with $K=16$ and learned position embeddings, and Pythia-410M with $K=16$ and rotary embeddings. We use the queries, keys and values of the first 100k held-out tokens (98 and 50 sequences, all 16 heads, 256 random queries per sequence). Since the key and value maps are already applied, we use Lemma~\ref{lem:boundary_distance} on the key and value paths themselves:
$|g_j-\widetilde g_j|\leq\|q\|_2\sqrt{(K-1)E^K_j}/\sqrt{d_k}$ and $\|u_j-\widetilde u_j\|_2\leq\sqrt{(K-1)E^V_j}$,
where $E^K_j$ and $E^V_j$ are the path energies of the keys and values in segment $j$. For Pythia we use the keys before rotation and add the rotary term of Appendix~\ref{app:rope}. Because $\|W(h_i-h_b)\|_2\leq\|W\|_{\mathrm{op}}\|h_i-h_b\|_2$, this version is never weaker than the one stated with $c_K$, $c_V$ and the path energy of $h$. Every inequality in the chain holds on all sampled queries.

Table~\ref{tab:prop4_check} shows that the low-energy condition does not hold. Inside a segment the values are as rough as a random reordering of the same tokens, and the keys are only somewhat smoother in \texttt{gpt2-medium}. The logit radius implied by the path energy is about 150 nats, so $\tanh$ saturates and the bound exceeds $2V_{\max}$ for every query. With the measured $\Delta_s$ and $\Delta_v$ the bound is tighter but still vacuous for almost all queries, and it does not rank queries by their error. The cause is that the logit gaps themselves are large: the measured $\Delta_s$ has median 10.8 and 12.6 nats, so $\tanh(\Delta_s)\approx1$ and the bound is close to $\Delta_v+2V_{\max}$ for almost every query, whatever the actual error. Across the 16 Pythia heads, the bound and the median error have rank correlation $-0.72$.

The error is governed instead by the attention mass that the mask removes. Let $m$ be the dense attention mass on keys outside the BBC pattern and $\bar u_M$ the attention-weighted mean of their values. Dense attention restricted to the visible keys equals $(1-m)\,o_{\mathrm{BBC}}$, so $o-o_{\mathrm{BBC}}=m(\bar u_M-o_{\mathrm{BBC}})$ exactly, and $m$ ranks queries by their error far better than the bound. In \texttt{gpt2-medium}, dense attention with the same $q$ and $k$ would put almost all of its mass outside the BBC pattern, spread over hundreds of positions. With rotary embeddings the logits stay local and $m$ is small. A trained BBC layer can therefore rely on the mask for locality; it does not have to approximate a dense layer. We read Proposition~\ref{prop:boundary_attention} as a qualitative continuity result, not as a quantitative estimate for trained models. The Pythia results are the same with $K=8$ and $K=32$.

\begin{table}[h]
\centering
\small
\caption{Proposition~\ref{prop:boundary_attention} at trained BBC layers (layer~2, 1B tokens). Medians over head--query pairs (401k for \texttt{gpt2-medium}, 205k for Pythia-410M). Path energies are divided by their expected value when the segment is put in random order, so 1 means as rough as random order. Rank correlations are Spearman correlations within each head, median over the 16 heads.}
\label{tab:prop4_check}
\begin{tabular}{lcc}
\toprule
 & \texttt{gpt2-medium} & Pythia-410M \\
 & {\footnotesize $K=16$, learned positions} & {\footnotesize $K=16$, rotary} \\
\midrule
Error of multiplicity-corrected attention, $\|o-\widetilde o\|_2/\|o\|_2$ & 1.70 & 0.14 \\
Key path energy relative to a random order of the segment & 0.57 & 0.97 \\
Value path energy relative to a random order of the segment & 1.00 & 0.97 \\
Logit radius from the path energy (nats) & 146 & 159 \\
Measured $\Delta_s$ (nats) & 10.8 & 12.6 \\
Bound / error, measured $\Delta_s,\Delta_v$ & 8.8 & 57 \\
Queries where the bound beats $\|o\|_2+\|\widetilde o\|_2$ & 0.003\% & 0.03\% \\
Rank correlation of bound and error, within head & 0.11 & $-0.02$ \\
Dense attention mass on removed keys, $m$ & 0.995 & 0.089 \\
Rank correlation of $m$ and BBC error, within head & 0.46 & 0.94 \\
\bottomrule
\end{tabular}
\end{table}

\section{Additional Theoretical Results}
\label{app:cc_properties}

\subsection{Scale and normalized roughness}
\label{app:cc_scale_norm}
Let the numerical stabilizer satisfy $\epsilon>0$.

Define
\[
\bar h=\frac1N\sum_{i=1}^N h_i,
\qquad
H_c=H-\mathbf1\bar h^\top.
\]
Since $L\mathbf1=0$, we have $E(H)=E(H_c)$.
Let
\[
S(H)=\|H_c\|_F^2,
\qquad
R_\epsilon(H)=\frac{E(H)}{S(H)+\epsilon}.
\]
Then
\begin{equation}
    E(H)=(S(H)+\epsilon)R_\epsilon(H).
\end{equation}
Raw energy therefore depends on both positional magnitude
and normalized roughness. For $S(H)>0$, the unstabilized
ratio $R(H)=E(H)/S(H)$ is invariant to uniform rescaling.

Temporarily treating $H$ as an independent optimization
variable, gradient flow for $\beta E(H)/(N-1)$ is
\begin{equation}
    \frac{dH}{dt}=-\frac{2\beta}{N-1}LH.
\end{equation}
It preserves the mean and gives
\begin{equation}
    \frac{d}{dt}\|H_c\|_F^2
    =
    -\frac{4\beta}{N-1}E(H)
    \leq0.
    \label{eq:cc-grad}
\end{equation}
This is a representation-space calculation. Parameter
updates act through the Jacobian of $H$ and need not
produce this flow, even when optimizing CC alone.
The calculation motivates separate scale and roughness
diagnostics rather than a monotonicity claim about
Transformer training.

\subsection{Finite-range graph energies}
\label{app:cc_k_connect}

Define
\[
E_r(H)=\sum_{i=1}^{N-r}\|h_{i+r}-h_i\|_2^2,
\qquad E_1(H)=E(H).
\]

\begin{proposition}[Adjacent path controls finite-range variation]
\label{prop:path_finite_range}
For $1\leq r<N$,
\[
E_r(H)\leq r^2E_1(H).
\]
For $1\leq k<N$, $\alpha_r\geq0$, $\alpha_1=1$, and
$E_{\leq k}^{\alpha}(H)=\sum_{r=1}^k\alpha_rE_r(H)$,
\begin{equation}
E_1(H)
\leq E_{\leq k}^{\alpha}(H)
\leq
\left(\sum_{r=1}^k\alpha_rr^2\right)E_1(H).
\end{equation}
\end{proposition}

\begin{proof}
Let $d_t=h_{t+1}-h_t$. Since
$h_{i+r}-h_i=\sum_{t=0}^{r-1}d_{i+t}$,
\[
\|h_{i+r}-h_i\|_2^2
\leq r\sum_{t=0}^{r-1}\|d_{i+t}\|_2^2.
\]
Each adjacent difference occurs in at most $r$ windows,
giving $E_r\leq r^2E_1$. The weighted bounds follow by
including the adjacent term and summing the upper bounds.
\end{proof}

Fixing $\alpha_1=1$ removes arbitrary overall scaling.
Within this family the path is the weakest unnormalized
penalty, while controlling finite-distance variation.
This ordering does not imply an ordering of validation NLL.

A nonconstant $r$-periodic sequence can satisfy $E_r=0$
but $E_1>0$, so a pure distance-$r$ graph with $r>1$
admits no uniform reverse bound.

For $\overline E_r=E_r/(N-r)$,
\[
\overline E_r(H)
\leq
r^2\frac{N-1}{N-r}\overline E_1(H).
\]
Comparisons between graph losses must therefore specify
their weights, normalization, and regularization coefficients.

\paragraph{The implemented CC-2 penalty.}

The experimental CC-2 objective combines differences before squaring.
For $N\geq3$, define its unnormalized energy by
\begin{equation}
    F_2(H)
    =
    \sum_{i=1}^{N-2}
    \|2h_{i+2}-h_{i+1}-h_i\|_2^2
    =
    (N-2)\mathcal{L}_{\rm CC\text{-}2}.
    \label{eq:cc2_residual_energy}
\end{equation}
Here $F_2$ differs from the distance-two edge energy $E_2$
defined above.

This objective has a directed Laplacian interpretation.
Orient unit-weight edges from each position $j$ to its
available predecessors $j-1$ and $j-2$. In the out-degree
Laplacian $L_{\leftarrow,2}$, row $j\geq3$ has coefficients
$2,-1,-1$ at columns $j,j-1,j-2$, respectively.
Let $B_2\in\mathbb{R}^{(N-2)\times N}$ contain rows
$3,\ldots,N$ of this Laplacian. Then
\begin{equation}
    F_2(H)
    =
    \|B_2H\|_F^2
    =
    \operatorname{Tr}(H^\top B_2^\top B_2H).
\end{equation}
Thus the implementation penalizes a squared Laplacian
residual over positions with two predecessors, rather
than a sum of separately squared edge differences.

\begin{proposition}[Path comparison for the implemented CC-2 penalty]
\label{prop:cc2_comparison}
For $N\geq3$,
\begin{equation}
\begin{aligned}
    F_2(H)
    &=E_1(H)+2E_2(H)\\
    &\quad
    -2\|h_2-h_1\|_2^2
    +\|h_N-h_{N-1}\|_2^2.
\end{aligned}
    \label{eq:cc2_edge_identity}
\end{equation}
Consequently,
\begin{equation}
    E_1(H)
    \leq F_2(H)+2\|h_2-h_1\|_2^2
    \leq 9E_1(H).
    \label{eq:cc2_path_comparison}
\end{equation}
In particular, $F_2(H)\leq9E_1(H)$.
\end{proposition}

\begin{proof}
Let $d_i=h_{i+1}-h_i$. Then
$2h_{i+2}-h_{i+1}-h_i=d_i+2d_{i+1}$, and
\begin{equation*}
\begin{aligned}
    \|d_i+2d_{i+1}\|_2^2
    &=
    2\|d_i+d_{i+1}\|_2^2\\
    &\quad-\|d_i\|_2^2+2\|d_{i+1}\|_2^2.
\end{aligned}
\end{equation*}
Summing over $i=1,\ldots,N-2$ gives
Equation~\ref{eq:cc2_edge_identity}.
Since $E_2(H)$ and the final-edge term are nonnegative,
this identity gives the lower bound in
Equation~\ref{eq:cc2_path_comparison}.

For the upper bound, Cauchy--Schwarz gives
\[
    \|d_i+2d_{i+1}\|_2^2
    \leq3\|d_i\|_2^2+6\|d_{i+1}\|_2^2.
\]
After summing,
\[
    F_2(H)
    \leq9E_1(H)-6\|d_1\|_2^2-3\|d_{N-1}\|_2^2.
\]
Adding $2\|d_1\|_2^2$ proves the upper bound.
\end{proof}

The initial-edge term cannot generally be omitted.
For example, the scalar trajectory $(h_1,h_2,h_3)=(0,2,1)$
has $F_2(H)=0$ but $E_1(H)=5$. More generally, nonzero
increments satisfying $d_{i+1}=-d_i/2$ give $F_2(H)=0$
while retaining positive path energy.

For the normalized training losses, the comparison becomes
\begin{equation}
\begin{aligned}
    \mathcal{L}_{\rm CC\text{-}2}
    &\leq
    9\frac{N-1}{N-2}\mathcal{L}_{\rm CC},\\
    \mathcal{L}_{\rm CC}
    &\leq
    \frac{N-2}{N-1}\mathcal{L}_{\rm CC\text{-}2}
    +\frac{2}{N-1}\|h_2-h_1\|_2^2.
\end{aligned}
    \label{eq:cc2_normalized_comparison}
\end{equation}
The initial-edge term is part of the theoretical comparison;
it is not added to the implemented loss. Adjacent CC
therefore controls the implemented CC-2 penalty, while
the reverse comparison requires control of the first
adjacent difference. The unconditional lower bound for
the pairwise graph family does not carry over to CC-2.

\subsection{Summed token-to-boundary distortion}

A token $r$ edges before its boundary satisfies
$\|h_i-h_{b_j}\|_2^2\leq rE_j$. Since
\[
\sum_{r=0}^{K-1}r=\frac{K(K-1)}2,
\qquad
\sum_jE_j\leq E(H),
\]
we obtain
\begin{equation}
    \sum_j\sum_{i\in I_j}\|h_i-h_{b_j}\|_2^2
    \leq
    \frac{K(K-1)}2E(H).
    \label{eq:boundary-prototype}
\end{equation}
This is a geometric consequence of low path energy.
Section~\ref{sec:cc_to_bbc} gives the corresponding
representative-attention approximation.

\subsection{CC and cosine distance}
\label{app:cos}

For this identity, assume both neighboring states are nonzero and omit
the numerical stabilizer $\varepsilon$. Define
\[
    \cos\theta_i
    =
    \frac{
        h_{i+1}^{\top}h_i
    }{
        \lVert h_{i+1}\rVert_2
        \lVert h_i\rVert_2
    },
\]
we have
\begin{equation}
    \frac{
        \lVert h_{i+1}-h_i\rVert_2^2
    }{
        \lVert h_{i+1}\rVert_2^2+\lVert h_i\rVert_2^2
    }
    =
    1
    -
    \frac{
        2\lVert h_{i+1}\rVert_2\lVert h_i\rVert_2
    }{
        \lVert h_{i+1}\rVert_2^2+\lVert h_i\rVert_2^2
    }
    \cos\theta_i .
\end{equation}
When neighboring hidden states have equal norm, this reduces to
$1-\cos\theta_i$. Thus normalized roughness is closely related to cosine
distance, while additionally accounting for differences in neighboring
hidden-state magnitudes.
\subsection{Proofs for Section~\ref{sec:bbc}}
\label{app:bbc_proofs}

\begin{proof}[Proof of Proposition~\ref{prop:bbc_structure_residual}]
(i) By Equation~\ref{eq:bbc_mask}, the query attends to $s+r+1$ of the
$t=sK+r+1$ causally visible positions, so $s(K-1)$ are removed: the $K-1$
non-boundary positions of each of the $s$ completed segments.
(ii) By (i), the mask equals $\{1,\ldots,t\}$ exactly when $s(K-1)=0$.
If $K=1$, this holds for every $t$. If $K\geq N$, then $t-1<K$, so $s=0$
for every $t\leq N$. If $1<K<N$, the query at $t=K+1\leq N$ has $s=1$ and
loses $K-1>0$ positions.
(iii) Because $a=\phi(S)$ is determined by $S$,
$p_\theta(S\mid C_s)=p_\theta(a\mid C_s)\,p_\theta(S\mid a,C_s)$; taking
negative logarithms gives Equation~\ref{eq:bbc_prob_residual}. The
counting bound follows from Equation~\ref{eq:bbc_residual_bound}.
\end{proof}

Summing (i) over positions, when $K$ divides $N$ the BBC layer removes
$N(K-1)(N-K)/(2K)$ of the $N(N+1)/2$ causal query--key pairs. This count
vanishes exactly at $K=1$ and $K=N$, is unchanged under $K\mapsto N/K$,
and, as a function of real $K$, is largest at $K=\sqrt N$. Per segment,
the realization-index bound grows linearly in $K$, while the number of
segments $N/K$ shrinks; summed over a sequence it stays near
$N\log_2|\mathcal{V}|$ for every $K$. The counting bound alone therefore
does not select $K$, nor does it specify which information the learned
carrier retains.

\subsection{What $K$ removes}
\label{app:bbc_theory}

The pair count ignores where attention goes. For a query $t$ and a key
$i\leq t$, write $\delta=t-i$, and $t=sK+r+1$ as in
Proposition~\ref{prop:bbc_structure_residual}.

\begin{lemma}[What BBC removes]
\label{lem:bbc_profile}
The key at distance $\delta$ is outside $\mathcal{A}_{\mathrm{BBC}}(t)$ if
and only if $\delta>r$ and $\delta\not\equiv r+1\pmod K$. If $sK\geq\delta$
and the offset $r$ is uniform on $\{0,\ldots,K-1\}$, the key is removed with
probability $\mu_K(\delta)=\max\{0,\min(\delta,K)-1\}/K$.
\end{lemma}

\begin{proof}
If $\delta\leq r$, then $t-\delta\geq sK+1$ is in the current segment and is
kept. Otherwise $1\leq t-\delta\leq sK$, and the key is kept if and only if
$K$ divides $t-\delta$, that is, $\delta\equiv t\equiv r+1\pmod K$. For
uniform $r$: if $\delta=0$, nothing is removed; if $1\leq\delta\leq K$, the
offsets with $r<\delta$ are $r=0,\ldots,\delta-1$, and only $r=\delta-1$
keeps the key, so $\delta-1$ of the $K$ offsets remove it; if $\delta>K$,
every offset has $r<\delta$ and exactly one keeps the key, so $K-1$ of the
$K$ offsets remove it.
\end{proof}

If a head puts mean weight $\pi(\delta)$ at distance $\delta$, independent of
the offset, the expected attention mass that the mask deletes is
$m(K)=\sum_\delta\pi(\delta)\mu_K(\delta)$. Unlike the pair count, it differs
for $K$ and $N/K$: a head that attends within $w\leq K$ tokens loses
$\sum_\delta\pi(\delta)(\delta-1)_+/K$, which falls like $1/K$. For one query
with dense weights $a_i$, values $v_i$, removed keys $M$ and
$m=\sum_{i\in M}a_i$, the BBC output
$o_{\mathrm{BBC}}=\sum_{i\notin M}a_iv_i/(1-m)$ satisfies
$o-o_{\mathrm{BBC}}=m(\bar u_M-o_{\mathrm{BBC}})$ with
$\bar u_M=\sum_{i\in M}a_iv_i/m$, so $m$ is exactly what the mask changes in a
fixed layer.

Table~\ref{tab:bbc_mass} measures $m$ on our 100M-token vanilla checkpoints.
It does not behave like a cost of BBC. At layer~2 the mask deletes most of the
vanilla layer's attention for every $K$ from 4 to 64, and in both models the
three values of $K$ with the best NLL delete the most. Across the swept
layers, the rank correlation between $m_l(8)$ and the BBC NLL of
Table~\ref{bbc-layers} is $-0.43$ for \texttt{gpt2-small} and $-0.08$ for
\texttt{gpt2-medium}. The same-density sliding
window, which gives each query as many keys as BBC but the nearest ones,
deletes less at layer~2 with $K=8$ ($0.57$ vs.\ $0.83$ for \texttt{gpt2-small},
$0.68$ vs.\ $0.83$ for \texttt{gpt2-medium}), yet on \texttt{gpt2-small} it
does not improve NLL (Table~\ref{tab:bbc-ablations}). A trained BBC layer therefore does not
approximate the vanilla layer; it uses the mask to restructure access, in line
with Appendix~\ref{app:prop4_check}.

\begin{table}[h]
\centering
\caption{Attention mass $m_2(K)$ that the BBC mask deletes at layer~2 of our 100M-token vanilla checkpoints (non-sink keys, mean over heads and over queries at positions $\geq64$, 16 sequences), next to the NLL of BBC at layer~2 (Table~\ref{bbc-layers} for $K\leq16$, Table~\ref{tab:bbc_k_sweep} for $K\geq32$).}
\label{tab:bbc_mass}
\small
\begin{tabular}{ccccc}
\toprule
 & \multicolumn{2}{c}{\texttt{gpt2-small}} & \multicolumn{2}{c}{\texttt{gpt2-medium}}\\
$K$ & $m_2(K)$ & NLL & $m_2(K)$ & NLL\\
\midrule
4 & 0.72 & 4.116 & 0.71 & 4.095\\
8 & 0.83 & 4.099 & 0.83 & 4.069\\
16 & 0.85 & 4.102 & 0.86 & 4.066\\
32 & 0.81 & 4.094 & 0.84 & 4.067\\
64 & 0.72 & 4.106 & 0.78 & 4.078\\
128 & 0.55 & 4.113 & 0.65 & 4.084\\
256 & 0.36 & 4.125 & 0.46 & 4.103\\
\bottomrule
\end{tabular}
\end{table}

\subsection{The CC weight $\beta$}
\label{app:cc_beta}

\begin{lemma}[CC energy is monotone in $\beta$]
\label{lem:cc_beta_monotone}
Fix the model class and the training distribution. For $\beta\geq0$, let $\theta_\beta$ minimize $\mathcal{L}_{\mathrm{NTP}}+\beta\mathcal{L}_{\mathrm{CC}}$.
For $0\leq\beta_1<\beta_2$,
$\mathcal{L}_{\mathrm{CC}}(\theta_{\beta_2})\leq\mathcal{L}_{\mathrm{CC}}(\theta_{\beta_1})$.
Moreover, $\theta_\beta$ minimizes $\mathcal{L}_{\mathrm{NTP}}$ among all models with $\mathcal{L}_{\mathrm{CC}}\leq\mathcal{L}_{\mathrm{CC}}(\theta_\beta)$.
\end{lemma}

\begin{proof}
Write $f=\mathcal{L}_{\mathrm{NTP}}$, $g=\mathcal{L}_{\mathrm{CC}}$ and
$\theta_k=\theta_{\beta_k}$. Optimality gives
$f(\theta_1)+\beta_1g(\theta_1)\leq f(\theta_2)+\beta_1g(\theta_2)$ and
$f(\theta_2)+\beta_2g(\theta_2)\leq f(\theta_1)+\beta_2g(\theta_1)$.
Adding them yields $(\beta_2-\beta_1)\bigl(g(\theta_2)-g(\theta_1)\bigr)\leq0$,
so $g(\theta_2)\leq g(\theta_1)$. If a model $\theta$ had
$g(\theta)\leq g(\theta_\beta)$ and $f(\theta)<f(\theta_\beta)$, its
penalized objective would be smaller than that of $\theta_\beta$.
\end{proof}

The lemma concerns exact minimizers, which training does not reach. Lower energy can also come from smaller scale, since $E(\alpha H)=\alpha^2E(H)$ while $R(\alpha H)=R(H)$ (Appendix~\ref{app:cc_scale_norm}), so the lemma alone does not show that $\beta$ shifts the balance between structure and residual. The following model makes this dependence explicit; it is graph Tikhonov smoothing \citep[Example~2]{shuman2013graphs} applied to hidden-state
trajectories.

\begin{proposition}[CC as a low-pass filter]
\label{prop:cc_filter}
Treat $H$ as a free variable, as in Appendix~\ref{app:cc_scale_norm}. Fix
$H_0\in\mathbb{R}^{N\times d}$ that is not constant across positions, and let
$\mu>0$. For $\beta\geq0$, the unique minimizer of
$\frac{\mu}{2}\|H-H_0\|_F^2+\frac{\beta}{N-1}E(H)$ is
\[
H^\star(\beta)=(I+\tau L)^{-1}H_0,\qquad\tau=\frac{2\beta}{\mu(N-1)},
\]
so the $r$-th graph-frequency component of $H_0$ is scaled by
$1/(1+\tau\lambda_r)$. For $1\leq m<N$, the centered residual share
$\rho^{c}_m(\beta)=\|R_m(H^\star(\beta))\|_F^2/\|H^\star(\beta)_c\|_F^2$ is non-increasing in $\beta$. It is strictly decreasing if $H_0$ has nonzero components both in modes $2,\ldots,m$ and in modes $m+1,\ldots,N$.
\end{proposition}

\begin{proof}
The objective is strictly convex with gradient
$\mu(H-H_0)+\frac{2\beta}{N-1}LH$, which vanishes exactly at
$H^\star(\beta)$. Let $a_r=\|(Q^\top H_0)_r\|_2^2$. Since $q_1$ is constant and
$\lambda_1=0$,
\[
\rho^{c}_m=\frac{A}{A+B},\qquad
A=\sum_{r=m+1}^{N}\frac{a_r}{(1+\tau\lambda_r)^2},\qquad
B=\sum_{r=2}^{m}\frac{a_r}{(1+\tau\lambda_r)^2}.
\]
Each nonzero term satisfies
$\frac{d}{d\tau}\log\frac{a_r}{(1+\tau\lambda_r)^2}=-2\varphi(\lambda_r)$ with
$\varphi(\lambda)=\lambda/(1+\tau\lambda)$, which is increasing in $\lambda$.
If $A,B>0$, then $\frac{d}{d\tau}\log A\leq-2\varphi(\lambda_{m+1})$ and
$\frac{d}{d\tau}\log B\geq-2\varphi(\lambda_m)$. The path-Laplacian eigenvalues
$\lambda_r=4\sin^2\bigl((r-1)\pi/(2N)\bigr)$ are distinct, so
$\lambda_{m+1}>\lambda_m$, and $A/B$, hence $\rho^{c}_m$, strictly decreases in $\tau$, which is proportional to $\beta$. If $A=0$ or $B=0$, $\rho^{c}_m$
is constant.
\end{proof}

The model inherits the limits of Appendix~\ref{app:cc_scale_norm}. It ignores causality, since $(I+\tau L)^{-1}$ mixes later positions into earlier ones, and its quadratic term charges for uniform rescaling, which rules out the scale route by assumption. With a feature-dependent curvature
$\frac12\operatorname{Tr}\bigl((H-H_0)W(H-H_0)^\top\bigr)$, $W\succ0$, the conclusion holds within each eigendirection of $W$ but can fail for the total share. Whether the centered residual share of a trained model falls with $\beta$ is therefore an empirical question.

\subsection{What $\beta$ controls}
\label{app:cc_theory}

Let $a_r=\|q_r^\top H\|_2^2$ and, for $r\geq2$, $p_r=a_r/S$, the spectral
distribution of $H_c$; write $\mathbb{E}_p$ and $\mathrm{Var}_p$ for moments of
$\lambda_r$ under $p$. Then $R=E/S=\mathbb{E}_p[\lambda]$.

\begin{proposition}[Scale and shape parts of the CC gradient]
\label{prop:cc_geom}
For non-constant $H$, $G=LH$ satisfies $\langle G,H_c\rangle_F=E$ and
$G=R\,H_c+G_\perp$ with $\langle G_\perp,H_c\rangle_F=0$,
$\|R\,H_c\|_F=R\sqrt S$, $\|G_\perp\|_F^2=S\,\mathrm{Var}_p(\lambda)$ and
$\|G\|_F^2=S\,\mathbb{E}_p[\lambda^2]$. Hence $R\sqrt S\leq\|G\|_F\leq2\sqrt{RS}$,
and $\nabla_H\mathcal{L}_{\mathrm{CC}}=2G/(N-1)$.
\end{proposition}

\begin{proof}
Since $L\mathbf1=0$, $G=LH_c=\sum_{r\geq2}\lambda_rq_rq_r^\top H$, so
$\langle G,H_c\rangle_F=\sum_r\lambda_ra_r=E$ and
$\|G\|_F^2=\sum_r\lambda_r^2a_r$. The component of $G$ along $H_c$ has
coefficient $E/S=R$, and Pythagoras gives the rest. Finally
$R^2\leq\mathbb{E}_p[\lambda^2]\leq\lambda_N\mathbb{E}_p[\lambda]\leq4R$.
\end{proof}

A gradient step on $E$ shrinks $H_c$ along itself, which changes $S$ and keeps
$R$, and moves orthogonally to $H_c$, which changes the shape and keeps $S$ to
first order. The ratio of the two parts is the coefficient of variation of
$\lambda$ under $p$. A penalty with $f(\alpha H)=f(H)$, such as normalized
roughness or cosine distance, satisfies $\langle\nabla f,H\rangle_F=0$ and has
only a shape part; $\mathcal{L}_{\mathrm{scale}}$ has only a scale part. In
Section~\ref{sec:experiments} and Appendix~\ref{app:cc-ablations}, scale alone recovers most of CC's gain at 100M tokens ($0.031$ of $0.036$ in NLL) but less than half at 300M tokens ($0.004$ of $0.010$), where CC is significantly better; shape alone recovers none.

\begin{proposition}[$\beta$ is relative to the layer's scale]
\label{prop:cc_scale}
Suppose $\mathcal{L}_{\mathrm{NTP}}(\alpha H)=\mathcal{L}_{\mathrm{NTP}}(H)$ for
all $\alpha>0$, as when $H$ is read only through per-token normalization. Then
$\nabla\mathcal{L}_{\mathrm{NTP}}(\alpha H)=\alpha^{-1}\nabla\mathcal{L}_{\mathrm{NTP}}(H)$
and $\nabla\mathcal{L}_{\mathrm{CC}}(\alpha H)=\alpha\nabla\mathcal{L}_{\mathrm{CC}}(H)$,
so $\rho=\|\nabla_H\mathcal{L}_{\mathrm{CC}}\|_F/\|\nabla_H\mathcal{L}_{\mathrm{NTP}}\|_F$
satisfies $\rho(\alpha H)=\alpha^2\rho(H)$. If a reparametrization $\theta\mapsto\theta_\alpha$
rescales $H$ by $\alpha$ and leaves $\mathcal{L}_{\mathrm{NTP}}$ unchanged, then
$\mathcal{L}_{\mathrm{NTP}}(\theta_\alpha)+\alpha^{-2}\beta\mathcal{L}_{\mathrm{CC}}(\theta_\alpha)
=\mathcal{L}_{\mathrm{NTP}}(\theta)+\beta\mathcal{L}_{\mathrm{CC}}(\theta)$.
\end{proposition}

\begin{proof}
Differentiate $\mathcal{L}_{\mathrm{NTP}}(\alpha H)=\mathcal{L}_{\mathrm{NTP}}(H)$
in $H$, and use that $\mathcal{L}_{\mathrm{CC}}$ is quadratic.
\end{proof}

In a pre-LN Transformer the residual stream is not read only through
normalization, so the invariance is approximate and we use the measured ratio
$\rho_l$ (Table~\ref{tab:cc_strength}). Adam's update is unchanged when the total
loss is multiplied by a constant, and to first order in $\beta$ the CC term
changes each coordinate's update in proportion to $\beta$ times its CC gradient
over the scale of its NTP gradient, which is why $\kappa_l=\beta\rho_l$ is the
natural strength.

\begin{table}[h]
\centering
\caption{CC strength by layer. $\rho_l$: ratio of the CC gradient to the NTP gradient at $\beta=1$ (code normalization) on our 100M-token vanilla checkpoints (16 sequences of 1024 tokens; standard errors below $4\%$). $\beta^\star$: best $\beta$ in Table~\ref{cc-layers}, the vertex of a parabola in $\log\beta$ through its three points, or the grid edge; for medium layer 22 and small layer 10 the parabola uses the best point of the three-seed sweep and its two neighbors (Table~\ref{tab:cc_beta_sweep}). Last column: the prediction $\kappa^\star/\rho_l$, with $\kappa^\star=0.00465$ the geometric mean of $\beta^\star\rho_l$ over medium layers 10--20.}
\label{tab:cc_strength}
\small
\begin{tabular}{lcccc}
\toprule
Layer & $\rho_l$ & $\beta^\star$ & $\beta^\star\rho_l$ & $\kappa^\star/\rho_l$\\
\midrule
\texttt{gpt2-medium} 22 & 3.54 & 0.0014 & 0.0049 & 0.0013\\
\texttt{gpt2-medium} 20 & 2.62 & 0.0018 & 0.0046 & 0.0018\\
\texttt{gpt2-medium} 18 & 2.03 & 0.0029 & 0.0059 & 0.0023\\
\texttt{gpt2-medium} 16 & 1.56 & 0.0030 & 0.0047 & 0.0030\\
\texttt{gpt2-medium} 14 & 1.21 & 0.0039 & 0.0047 & 0.0039\\
\texttt{gpt2-medium} 12 & 0.99 & 0.0050 & 0.0049 & 0.0047\\
\texttt{gpt2-medium} 10 & 0.86 & 0.0039 & 0.0034 & 0.0054\\
\texttt{gpt2-medium} 8 & 0.84 & $\geq0.005$ & $\geq0.0042$ & 0.0055\\
\texttt{gpt2-medium} 6 & 0.83 & $\geq0.005$ & $\geq0.0042$ & 0.0056\\
\texttt{gpt2-medium} 4 & 0.83 & $\geq0.005$ & $\geq0.0042$ & 0.0056\\
\texttt{gpt2-medium} 2 & 0.83 & $\geq0.005$ & $\geq0.0042$ & 0.0056\\
\midrule
\texttt{gpt2-small} 10 & 2.20 & 0.0024 & 0.0053 & 0.0021\\
\texttt{gpt2-small} 8 & 1.37 & 0.0028 & 0.0038 & 0.0034\\
\texttt{gpt2-small} 6 & 0.79 & $\geq0.005$ & $\geq0.0039$ & 0.0059\\
\texttt{gpt2-small} 4 & 0.60 & 0.0038 & 0.0023 & 0.0077\\
\texttt{gpt2-small} 2 & 0.73 & $\geq0.005$ & $\geq0.0036$ & 0.0064\\
\bottomrule
\end{tabular}
\end{table}

Across \texttt{gpt2-medium} layers 10--20, $\rho_l$ grows $3.1\times$ and
$\beta^\star$ falls $2.2\times$, so $\beta^\star\rho_l$ stays within
$[0.0034,0.0059]$. Its geometric mean predicts the best $\beta$ at medium
layer 22 and small layer 10 to within $15\%$ of the three-seed vertices, and
at medium layers $\leq8$ it places the optimum just above the grid, where
Table~\ref{cc-layers} has its best value. On \texttt{gpt2-small} the rule
also holds at layers 2--8 except layer 4, which it misses by $2\times$; there
the three NLL values of Table~\ref{cc-layers} differ by at most $0.012$.

Lemma~\ref{lem:cc_beta_monotone} also bounds the price of CC. With $f$, $g$ and
$\theta_\beta$ as there and $\theta_0$ a minimizer of $f$,
\[
0\leq f(\theta_\beta)-f(\theta_0)\leq\beta\bigl(g(\theta_0)-g(\theta_\beta)\bigr)\leq\beta\,g(\theta_0),
\]
because $f(\theta_\beta)+\beta g(\theta_\beta)\leq f(\theta_0)+\beta g(\theta_0)$
and $f(\theta_0)\leq f(\theta_\beta)$. If $f,g$ are $C^2$ near a nondegenerate
$\theta_0$ and $\theta_\beta$ is the local branch, the implicit function theorem
gives $f(\theta_\beta)-f(\theta_0)=\frac{\beta^2}2\nabla g^\top(\nabla^2f)^{-1}\nabla g+o(\beta^2)$,
while $g(\theta_0)-g(\theta_\beta)=\beta\,\nabla g^\top(\nabla^2f)^{-1}\nabla g+o(\beta)$.
The fitting price is second order in $\beta$; the next proposition shows that
the statistical benefit is first order.

\begin{proposition}[Some $\beta>0$ helps, and $\beta^\star$ shrinks with data]
\label{prop:cc_mest}
Let the population excess loss be
$\mathcal{R}(\theta)=\frac12(\theta-\theta_*)^\top\mathsf F(\theta-\theta_*)$ with
$\mathsf F\succ0$, let $\hat\theta_0$ be an unbiased estimate of $\theta_*$ with
covariance $\mathsf V/n$, $\mathsf V\succ0$, and let
$\hat\theta_\beta=\arg\min_\theta\frac12(\theta-\hat\theta_0)^\top\mathsf F(\theta-\hat\theta_0)+\beta\Omega(\theta)$
with $\Omega(\theta)=\frac12\theta^\top\mathsf P\theta-c^\top\theta$,
$\mathsf P\succeq0$, $\mathsf P\neq0$. With $g_*=\nabla\Omega(\theta_*)$ and
$r(\beta)=\mathbb{E}\,\mathcal{R}(\hat\theta_\beta)$,
\[
r(\beta)=r(0)-\frac{\beta}{n}\operatorname{tr}(\mathsf P\mathsf V)
+\frac{\beta^2}{2}\Bigl(g_*^\top\mathsf F^{-1}g_*+\frac3n\operatorname{tr}(\mathsf P\mathsf F^{-1}\mathsf P\mathsf V)\Bigr)+O(\beta^3),
\]
so $r'(0)=-\operatorname{tr}(\mathsf P\mathsf V)/n<0$. With $\beta=b/n$,
$n^2\bigl(r(b/n)-r(0)\bigr)\to-b\operatorname{tr}(\mathsf P\mathsf V)+\frac{b^2}2g_*^\top\mathsf F^{-1}g_*$,
which is minimized at $b^\star=\operatorname{tr}(\mathsf P\mathsf V)/(g_*^\top\mathsf F^{-1}g_*)$.
\end{proposition}

\begin{proof}
$\hat\theta_\beta-\theta_*=A_\beta(\hat\theta_0-\theta_*)-\beta(\mathsf F+\beta\mathsf P)^{-1}g_*$
with $A_\beta=(\mathsf F+\beta\mathsf P)^{-1}\mathsf F$. The cross term has zero
mean, so
$r(\beta)=\frac1{2n}\operatorname{tr}(A_\beta^\top\mathsf FA_\beta\mathsf V)+\frac{\beta^2}2g_*^\top(\mathsf F+\beta\mathsf P)^{-1}\mathsf F(\mathsf F+\beta\mathsf P)^{-1}g_*$.
With $B=\mathsf F^{-1}\mathsf P$, $A_\beta=I-\beta B+\beta^2B^2+O(\beta^3)$ and
$A_\beta^\top\mathsf FA_\beta=\mathsf F-2\beta\mathsf P+3\beta^2\mathsf P\mathsf F^{-1}\mathsf P+O(\beta^3)$.
Finally $\operatorname{tr}(\mathsf P\mathsf V)>0$ for $\mathsf P\succeq0$,
$\mathsf P\neq0$, $\mathsf V\succ0$.
\end{proof}

For CC, $\mathsf P$ is the path Laplacian acting on the representation,
$\mathsf V$ the covariance of the representation's estimation error, and $g_*$
the CC gradient at the population representation. The best weight is large when
the estimation error is rough and the target trajectory is smooth, it decays like
$1/n$, and the gain decays like $1/n^2$ against an excess loss of order $1/n$.
So CC's relative benefit should shrink with the token budget, and the best
$\beta$ should fall. Our runs agree in direction: CC adds about $0.3\%$ on top of
BBC at 1B tokens but $0.06\%$ at 20B ($0.14\%$ with adaptive dropout), and the 20B-horizon sweep selects a
smaller $\beta$. BBC's gain does not shrink with training, so BBC does not act as
a regularizer of this kind. The model gives the sign and the direction of the
scaling, not the constants.

\begin{proposition}[CC smoothing as structure estimation]
\label{prop:cc_bayes}
Per coordinate, let $x_{i+1}=x_i+\eta_i$ with $\eta_i\sim\mathcal{N}(0,\varsigma^2)$
and a flat prior on $x_1$, and let $h_i=x_i+w_i$ with
$w_i\sim\mathcal{N}(0,\sigma^2)$, all independent. Then
$X\mid H\sim\mathcal{N}\bigl((I+\tau L)^{-1}H,\ \sigma^2(I+\tau L)^{-1}\bigr)$ with
$\tau=\sigma^2/\varsigma^2$. The posterior mean squared error per position is
$\frac{\sigma^2}{N}\sum_r(1+\tau\lambda_r)^{-1}$, which tends to
$\sigma^2/\sqrt{1+4\tau}$ as $N\to\infty$, and
$\mathbb{E}(h_{i+k}-h_i)^2=2\sigma^2+k\varsigma^2$.
\end{proposition}

\begin{proof}
The log posterior is
$-\frac1{2\sigma^2}\|H-X\|^2-\frac1{2\varsigma^2}X^\top LX+\mathrm{const}$, with
precision $\sigma^{-2}(I+\tau L)$. The eigenvalues are
$\lambda_r=4\sin^2\bigl((r-1)\pi/(2N)\bigr)$, and the Riemann sum converges to
$\frac1\pi\int_0^\pi\frac{d\theta}{1+2\tau-2\tau\cos\theta}=(1+4\tau)^{-1/2}$.
\end{proof}

By Proposition~\ref{prop:cc_filter}, the minimizer of
$\frac\mu2\|H-H_0\|_F^2+\frac{\beta}{N-1}E(H)$ is $(I+\tau L)^{-1}H_0$ with
$\tau=2\beta/(\mu(N-1))$. Matching the two gives
$\beta=\frac12\mu(N-1)\,\sigma^2/\varsigma^2$: the CC weight is the NTP curvature
times the ratio of residual to structure variance, and the variogram identifies
that ratio from data. Under Proposition~\ref{prop:cc_scale}, $\mu$ scales like
$1/S$, so $\beta S$ is set by $\sigma^2/\varsigma^2$. The model treats the white
part as removable; at early layers it is token identity, which later layers
need, so the model fixes the meaning of $\beta$ but not the layer.

\subsection{Rotary position embeddings}
\label{app:rope}

With rotary position embeddings \citep{su2024roformer}, queries and keys are
rotated by position. Let $U_n\in\mathbb{R}^{d_k\times d_k}$ be block diagonal
with $2\times2$ rotation blocks of angles $n\omega_1,\ldots,n\omega_{d_k/2}$
($\omega_\ell=0$ for blocks left unrotated, as in partial rotary embeddings),
write $u^{(\ell)}$ for the $\ell$-th two-dimensional block of a vector $u$, and
let $\Omega$ scale the $\ell$-th block by $\omega_\ell$. The logit of a query
$q$ at position $t$ for a key $k$ at position $i$ is
$q^\top U_{i-t}k/\sqrt{d_k}$, and values are not rotated. Each block of
$U_n-I$ has operator norm $2|\sin(n\omega_\ell/2)|\leq\min\{2,|n|\omega_\ell\}$,
so for all vectors $u$ and $v$,
\begin{equation}
    |v^\top(U_n-I)u|\leq\sum_\ell\min\{2,|n|\omega_\ell\}\,\|v^{(\ell)}\|_2\|u^{(\ell)}\|_2,
    \qquad
    \|(U_n-I)u\|_2\leq\min\bigl\{2\|u\|_2,\ |n|\,\|\Omega u\|_2\bigr\}.
    \label{eq:rope_rotation}
\end{equation}

\textbf{Rotary version of Proposition~\ref{prop:boundary_attention}.}\ \
Proposition~\ref{prop:boundary_attention} holds for any logits; what changes
is the bound on $|g_j-\widetilde g_j|$ in Section~\ref{sec:cc_to_bbc}. For $i$
in a completed segment with boundary $b=b_j$ and keys $k_i=W_Kh_i$,
$U_{i-t}k_i-U_{b-t}k_b=U_{i-t}(k_i-k_b)+U_{b-t}(U_{i-b}-I)k_b$, and
$U_{b-t}^\top q$ has the same block norms as $q$.
Lemma~\ref{lem:boundary_distance} and Equation~\ref{eq:rope_rotation} give
\[
|g_j-\widetilde g_j|
\leq c_K\sqrt{(K-1)E_j}
+\frac{1}{\sqrt{d_k}}\sum_\ell\min\{2,(K-1)\omega_\ell\}\,
\|q^{(\ell)}\|_2\|k_b^{(\ell)}\|_2,
\]
while the value bound is unchanged, so Proposition~\ref{prop:boundary_attention}
applies with $\Delta_s$ at most the maximum of this bound over $j$. The second
term does not depend on $E_j$; it is small when the query and the boundary key
share little energy in fast-rotating blocks.

\textbf{Rotary version of Proposition~\ref{prop:bbc_smoothness}.}\ \ With
$\widetilde q_i=U_iq_i$ and $\widetilde k_b=U_bk_b$, the boundary logits are
$\widetilde q_i^\top\widetilde k_b/\sqrt{d_k}$, so
Proposition~\ref{prop:bbc_smoothness} holds with $Q^{\mathrm{qry}}$ and $K_B$
replaced by their rotated versions. Since
$\widetilde q_{i+1}-\widetilde q_i=U_{i+1}(q_{i+1}-q_i)+U_i(U_1-I)q_i$,
Minkowski's inequality and Equation~\ref{eq:rope_rotation} give
\[
\sqrt{E_{\mathrm{intra}}(\widetilde Q^{\mathrm{qry}})}
\leq\sqrt{E_{\mathrm{intra}}(Q^{\mathrm{qry}})}
+\Bigl(\sum_i\|\Omega q_i\|_2^2\Bigr)^{1/2},
\]
where $i$ runs over the first position of each adjacent pair in the segment,
and $\|\widetilde K_B\|_{\mathrm{op}}\leq\|K_B\|_F$ because rotations preserve
row norms. The extra term is nonzero even for constant queries.

\section{Issues in the STP/LLM-JEPA Implementation}
\label{app:implementation_issuses}

CC is partly inspired by the geometric prior on hidden-state trajectories introduced in \citep{huang2026stp}.
While reproducing STP and LLM-JEPA results, we identified a subtle issue in the official
implementation of both papers that affects the reported performance gains. The issue originates in the
\texttt{DataCollator} used for training, which masks the labels of all padding tokens. For models whose
\texttt{EOS} and \texttt{PAD} tokens share the same id, this also masks the \texttt{EOS} token itself, removing the
training signal that teaches the model when to terminate the generation. Models with distinct \texttt{EOS} and
\texttt{PAD} tokens are unaffected. We flagged this to the authors, who
confirmed the issue and have since updated the implementation\footnote{\href{https://github.com/galilai-group/llm-jepa/commit/1c07b44fa2b0408a064be0ffc7816b5281384d7f}{https://github.com/galilai-group/llm-jepa/commit/1c07b44fa2b0408a064be0ffc7816b5281384d7f}}.

After fixing the bug, we reproduced the results using \texttt{Llama-3.2-1B-Instruct} on the NL-RX-SYNTH dataset. NTP
accuracy rose sharply, from 62.35\% to 86.2\%, while STP improved more modestly, from 84.35\% to 86.31\%.
This gap suggests that STP is notably more robust to this failure mode than NTP. To
probe this further, we ran a preliminary experiment masking the last two tokens. After one
epoch, STP recovered to 72.55\% accuracy, compared to 25.4\% for NTP. This robustness did not extend
to masking more than two tokens, however. Understanding why STP is able
to partially recover from this kind of signal loss, while NTP is not, remains an interesting
direction for future work.

\end{document}